\documentclass[12pt]{article}
\usepackage{amsmath}
\usepackage{graphicx}
\usepackage{enumerate}
\usepackage{natbib}
\usepackage{url} 
\RequirePackage{amsmath,amsthm,amsfonts,amssymb,bm,mathrsfs}

\usepackage{dsfont,comment}
\usepackage{bm}
\usepackage{color,soul}
\usepackage{hyperref}
\usepackage{multirow}
\usepackage{natbib}
\usepackage{enumitem}
\usepackage{placeins}
\usepackage[ruled,linesnumbered]{algorithm2e}
\usepackage[normalem]{ulem}

\usepackage[utf8]{inputenc} 
\usepackage[T1]{fontenc}    
\usepackage{hyperref}       
\usepackage{url}            
\usepackage{booktabs}       
\usepackage{amsfonts}       
\usepackage{nicefrac}       
\usepackage{microtype}      
\usepackage{xcolor}         
\usepackage{amsmath}
\usepackage{amssymb}
\usepackage{mathtools}
\usepackage{amsthm}
\usepackage{bm}
\usepackage{placeins}

\usepackage{graphicx}
\usepackage{subfigure}
\usepackage[capitalize,noabbrev]{cleveref}

\theoremstyle{plain}
\newtheorem{theorem}{Theorem}[section]
\newtheorem{proposition}[theorem]{Proposition}
\newtheorem{lemma}[theorem]{Lemma}
\newtheorem{corollary}[theorem]{Corollary}
\theoremstyle{definition}
\newtheorem{definition}[theorem]{Definition}
\newtheorem{assumption}[theorem]{Assumption}
\newtheorem{example}[theorem]{Example}

\theoremstyle{remark}

\def\1{\bm{1}}

\usepackage{cleveref}
\usepackage{ulem}
\usepackage{graphicx}
\usepackage{xr}

\DeclareMathOperator*{\argmax}{arg\,max}
\DeclareMathOperator*{\argmin}{arg\,min}

\title{Learning Guarantee of Reward Modeling Using Deep Neural Networks}

\author{Yuanhang $\rm{Luo}^{a}$\thanks{Equally contributed authors; ${}^{\dagger}$Co-corresponding authors.},\hspace{0.2cm}
		Yeheng $\rm{Ge}^{a*}$,\hspace{0.2cm}
		Ruijian $\rm{Han}^{a\dagger}$,\hspace{0.2cm}
        and
		Guohao $\rm{Shen}^{b\dagger}$\hspace{0.2cm}
		\\
		{\small {\small {$\it^{a}$  Department of Data Science and Artificial Intelligence, The Hong Kong Polytechnic University  } }}\\
        {\small {\small {$\it^{b}$  Department of Applied Mathematics, The Hong Kong Polytechnic University  } }}
	}

\begin{document}

\maketitle

\begin{abstract}
In this work, we study the learning theory of reward modeling with pairwise comparison data using deep neural networks. We establish a novel non-asymptotic regret bound for deep reward estimators in a non-parametric setting, which depends explicitly on the network architecture. Furthermore, to underscore the critical importance of clear human beliefs, we introduce a margin-type condition that assumes the conditional winning probability of the optimal action in pairwise comparisons is significantly distanced from 1/2. This condition enables a sharper regret bound, which substantiates the empirical efficiency of Reinforcement Learning from Human Feedback and highlights clear human beliefs in its success. Notably, this improvement stems from high-quality pairwise comparison data implied by the margin-type condition, is independent of the specific estimators used, and thus applies to various learning algorithms and models.
\end{abstract}

\section{Introduction}
\label{introduction}

Reinforcement Learning from Human Feedback (RLHF) has proven highly effective in aligning large language models with human preferences and expert policies \citep{christiano2017deep}. A significant advancement in this domain is Direct Preference Optimization (DPO), which enhances RLHF by learning directly from pairwise comparison data, bypassing the need for an iterative reinforcement learning fine-tuning pipeline. This method not only improves efficiency but also strengthens alignment with human preferences \citep{rafailov2024direct}, particularly in contexts with intuitively clear feedback, such as recommendation systems and image generation.

Recent studies have explored the theoretical underpinnings of RLHF, yet performance guarantees remain an open challenge. \citet{zhu2023principled} investigated the theoretical properties of reward modeling based on action-based comparisons, while \citet{chen2022human} and \citet{saha2023dueling} focused on trajectory-based comparisons. However, these analyses have largely focused on specific functional classes with limited representational power, neglecting bias due to approximation errors under misspecification. Despite the success of deep neural networks (DNNs) in RLHF due to their expressive power, their approximation properties and estimation bias remain theoretically underexplored. This gap in the literature highlights the necessity for a more comprehensive theoretical framework to fully leverage the potential of DNN estimators in RLHF.

Sample efficiency is a fundamental theoretical challenge in RLHF. Unlike traditional RL, which often requires vast amounts of environmental interaction, RLHF can achieve strong performance with relatively few pairwise comparison data \citep{christiano2017deep}. This efficiency gain likely stems from the informational richness of human preferences: by explicitly ranking actions or outputs, human feedback provides a dense learning signal that reduces the need for exploration \citep{wang2024secrets}. However, this advantage depends critically on the quality of preferences. Ambiguous or conflicting feedback can obscure the true reward function, while high-consensus datasets, such as those used to fine-tune InstructGPT \citep{ouyang2022training}, enable robust reward modeling. Existing RLHF studies emphasize the empirical benefits of unambiguous feedback, but theoretical quantification of how such clarity translates to sample efficiency remains a critical gap.

To address these challenges, this paper offers non-parametric theoretical guarantees for DNN-based reward modeling and develops a mathematical model to calibrate clear preferences in RLHF, enhancing our understanding of its sample efficiency. Specifically, our contributions focus on the following aspects: 
\begin{itemize}
     \item  
    {\bf Regret Bounds of DNN Estimators.} We establish regret bounds for reward modeling utilizing deep neural networks with pairwise comparison data. These bounds balance approximation and stochastic error based on the sample size, providing critical insights into the efficacy of DNN-based estimators in RLHF and underscoring their sample efficiency.
 
     \item {\bf Calibration of Clear Human Beliefs.} We introduce a novel margin-type condition to calibrate clear human beliefs in RLHF. This condition suggests high-quality pairwise comparison datasets and reveals the structure of the underlying reward, leading to sharper regret bounds. Our findings emphasize the importance of clear human beliefs in the success of RLHF, with these theoretical improvements being applicable across diverse learning algorithms.

    \item {\bf Empirical Investigations.} We demonstrate the necessity for high-quality pairwise comparison data and validate our theory through experiments with varying DNN architectures, emphasizing the broad applicability of our theoretical results in reward modeling.

\end{itemize}

The paper is organized as follows: Section \ref{sec:comparison_model} introduces the pairwise comparison model and a margin-type condition that accelerates regret convergence. Section \ref{Theoretical_Analysis} derives architecture-dependent non-asymptotic regret bounds and their implications for deep reward estimators. Numerical experiments in Section \ref{sec:exp} examine how network architecture and comparison data quality affect estimation performance. Sections \ref{sec:related_work}-\ref{sec:conclusion} provide a literature review and conclusion. Technical details are deferred to the Appendix.

\section{Pairwise Comparison, Margin-type Condition and Sharper Bound}\label{sec:comparison_model}

In this work, we consider the reward modeling in the action-based pairwise comparison case.
Let $\mathcal{S}$ be the set of states (prompts), and $\mathcal{A}$ be the set of actions (responses). We consider a pairwise comparison dataset 
$\{s^{i},a^{i}_{1},a^{i}_{0},y^i\}^N_{i=1}$ with sample size $N$: the state $s^{i}$ is sampled from the probability measure on state space $\mathcal{S}$,  
denoted by $\rho_s$; Conditioning on the state $s^i$, the action pair $(a^{i}_1,a^{i}_0)$ is sampled from some joint distribution $\mathbb{P}(a_1,a_0|s^i)$; The comparison outcome $y^i$ indicates the preference between $a^{i}_{1}$ and $a^{i}_{0}$. Specifically, $a^{i}_{1}$ is preferred over $a^{i}_{0}$ if $y^i > 0$ and conversely, $a^{i}_{0}$ is preferred if $y^i < 0$.
It is worth noting that we do not restrict the outcome to a binary format, and it accommodates various types of outcomes discussed in the literature. For simplicity, readers may consider the binary case where $y^i$ takes on values of either $-1$ or $1$. We define the reward function $r: \mathcal{S} \times \mathcal{A} \rightarrow \mathbb{R}$, which evaluates the reward of taking each action at a given state. We denote $d$ as the dimension of the input for the reward function $r$. 

For any reward function $r$, we denote the decision maker as $\pi_{r}(s) = \argmax_{a  \in \mathcal{A}} \ r(s,a)$. Let $r^*(s,a)$ denote the underlying true reward function. We define the optimal action for the state $s$ by $\pi_{r^*}(s) = \argmax_{a  \in \mathcal{A}} \ r^*(s,a)$. For an estimated reward function $r$, we are interested in the regret of the induced $\pi_{r}(s)$, which is
\begin{equation}\label{loss_performance}\mathcal{E}(r)= \int_{\mathcal{S}} r^*(s,\pi_{r^*}(s)) - r^*(s,\pi_{r}(s))d\rho_{s}.
\end{equation} The regret (\ref{loss_performance}) is an intrinsic measure for evaluating RLHF
\citep{zhu2023principled, zhan2023provable}. 
It is important to note that the policy $\pi_{r}$ is induced from the reward  $r$ and the resulting regret is determined by $r$. 
In practice, we optimize the comparison model using crowd-sourced comparison outcomes. 
Even if the reward function is not perfectly estimated, there remains an opportunity to derive a correct policy and achieve low regret in reinforcement learning tasks. This potential stems from clear human beliefs, which suggest a significant gap between the reward of the optimal action and its alternatives. However, the critical role of reward differences between actions is often overlooked in pairwise comparison analysis, which typically relies on the smoothness of the reward function. In light of this, we are motivated to quantify the effects of reward differences on regret, capturing the reward gap between actions.

As discussed in \cite{wang2024secrets,song2024preference,zhan2023provable}, we model the relationship between comparison response and the difference of the rewards $r^*(s,a_1)-r^*(s,a_0)$. Specifically, the probability of the event that $a_1$ is preferred over $a_0$ under the state $s$ can be expressed as:
$$
\mathbb{P}(y>0 \mid s,a_1,a_0)=\int_0^{\infty} g(y , r^*(s,a_1)-r^*(s,a_0)) d y,
$$
where the function $g$ represents the probability density function of the comparison outcome $y$ and in this paper we consider a general parametrization for $g$.
It is worth noting that the success of RLHF is largely attributed to clear human preferences, while incorrect or ambiguous preference labels can lead to significant performance deterioration in practice 
\citep{saha2023dueling, wang2024secrets, chen2024extendingdirect}.
To calibrate the clear human preferences, we propose the following margin condition in the pairwise comparison dataset.
\begin{assumption}[Margin Condition for the Human Preference]
\label{noisep}
For any action pairs  $(\pi_{r^*}(s),a^\prime)$ where 
$a^\prime \in \mathcal{A} \setminus \pi_{r^*}(s)$ and $t \in (0,1/2)$, we have
\begin{equation*}
    \begin{aligned}
        &\mathbb{P}_{\mathcal{S}}\left(\mathbb{P}(y>0 \mid s,a_1=\pi_{r^*}(s),a_0=a^\prime)-\frac{1}{2}\leq t\right)\\
        &:=  \int_{\mathcal{S}} \1\left\{\int_{0}^{\infty} g(y, r^*(s,\pi_{r^*}(s))- r^*(s,a^\prime))dy -\frac{1}{2} \leq t \right\} d \rho_{s} \leq c_g t^{\frac{\alpha}{1-\alpha}},
    \end{aligned}
\end{equation*}
where $c_g>0$ is a universal constant and  {$\alpha \in (0,1)$} is the coefficients for quantifying the clear human belief.  The larger $\alpha$ indicates a clearer preference in the pairwise comparison dataset.
\end{assumption}
Assumption \ref{noisep} implies that experts have a clear tendency between the optimal action and the other for most states $s$, under which the winning probability of the optimal action is bounded away from 1/2. It is worth noting that $\alpha = 0$ and $\alpha = 1$ correspond to two extreme cases, respectively, for the case without any margin-type assumption and the noiseless case.
To better understand  Assumption \ref{noisep}, we take a closer look at its implication on the underlying reward function. 
Here, we present two classical comparison models with {$y\in \{-1,1\}$} as examples.

\begin{example}[Bradley-Terry (BT) model \citep{bradley1952rank}] \label{BT} 
The comparison function is
\begin{equation*}
g(y,u)=\mathbf{1}(y=1) \cdot \frac{\exp(u)}{1+\exp(u)}+\mathbf{1}(y=-1) \cdot \frac{\exp(-u)}{1+\exp(-u)}.
\end{equation*}
Given a particular state-action pair, the probability of observing the outcome $y>0$ is $\exp (r^*(s,a_1)-r^*(s,a_0))/(1+\exp(r^*(s,a_1)-r^*(s,a_0))).$
\end{example}

\begin{example}[Thurstonian model \citep{thurstone2017law}]
\label{thurstonian}
The comparison function is
    \begin{equation*}
g(y,u)
        = \mathbf{1}(y=1) \cdot \Phi(u)+\mathbf{1}(y=-1) \cdot (1-\Phi(u)),
\end{equation*}
where $\Phi(u)$ is the cumulative distribution function of the standard normal distribution.
Then we have $\mathbb{P}(y > 0 \mid s,a_1,a_0) = \Phi(r^*(s,a_1)-r^*(s,a_0))$.
\end{example}

The \textit{BT model} and  \textit{Thurstonian model} are widely considered in reward modeling \citep{christiano2017deep,rafailov2024direct,siththaranjan2023distributional}.
In addition, some other comparison models are also adopted. For example, the Rao-Kupper model \citep{rao1967ties} and the Davidson model \citep{davidson1970extending} are used to tackle pairwise comparisons with ties (abstentions), where $y$ takes values from $\{-1, 0,1\}$ \citep{chen2024extendingdirect}. 
Notably, the general comparison framework discussed in this work encompasses all such examples.

The comparison model connects the underlying reward function to human preferences within the observed comparison datasets. By considering the clear preference data outlined in Assumption  \ref{noisep}, we could further reveal the specific structure of the reward function, which is summarized in the following Lemma \ref{noiser}.

\begin{lemma}
\label{noiser}
    Given Assumption \ref{noisep}, with $\alpha \in (0,1)$ and $t \in (0, c_{r^*})$ {where $c_{r^*}$ is the upper bound of the true reward defined in Assumption \ref{dynamic_range}}, there exists a universal constant  $c_g^\prime$ such that 
    \begin{align*}
        \int_{\mathcal{S}} \1\left\{r^*(s,\pi_{r^*}(s))-\max_{a\in \mathcal{A}\setminus \pi_{r^*}(s)}r^*(s,a) \leq t\right\} d \rho_{s} \leq c'_g t^{\frac{\alpha}{1-\alpha}}.
    \end{align*}
Specifically, $c^\prime_g= (1/4)^{\alpha/(1-\alpha)}c_g$ in the BT model and $c^\prime_g =  (1/2\pi)^{\alpha/(2-2\alpha)}c_g$ in the Thurstonian model where $c_g$ is a universal constant presented in Assumption \ref{noisep}.
\end{lemma}
Lemma \ref{noiser} implies that a clear preference in the comparison dataset is determined by the reward margin between the two actions. 
Its validity also depends on the properties of the comparison function specified in Definition \ref{comparison_function}, making it applicable to the general comparison model.
Unlike the existing literature, which imposes unverifiable conditions directly on the reward structure, Lemma \ref{noiser} is derived from Assumption \ref{noisep} on high-quality preference datasets. This approach is more valid and enjoys greater generalization ability compared to existing conditions \citep{Kim2021Fast, shi2023value,zhan2023provable}.

\subsection{Sharper Regret Bound}

In this section, we present the regret bounds of the decision maker $\pi_{r}(s)$ defined in (\ref{loss_performance}) with and without Assumption \ref{noisep}.

\begin{theorem}[Faster Rate with Margin Condition]
\label{faster}
Let $r$ be some reward function estimator. With Assumption \ref{noisep} holds and the margin parameter $\alpha \in (0,1)$, there exists a universal constant $c_1>0$, such that
$$
\mathcal{E}(r)\leq c_1\left(\|r-r^*\|^2_{L^2(\mathcal{S}, \ell^2)}\right)^{\frac{1}{3-2\alpha}},
$$
where the norm $\|\cdot\|_{L^2(\mathcal{S}, \ell^2)}$ is defined in (\ref{L2_distance}). 
\end{theorem}
Theorem \ref{faster} demonstrates that clear preferences, quantified by $\alpha\in (0,1)$, lead to improved regret bounds in reward modeling \citep{zhu2023principled}. When a hard margin is imposed, that is, $\alpha \rightarrow 1$, the regret of the ``greedy'' policy induced by the estimated reward function is of the order $\mathcal{O}(\|r-r^*\|^2_{L^2(\mathcal{S}, \ell^2)})$.

\begin{corollary}[Regret Bound without Margin Condition]
\label{without_TNC}
Let $r$ be some reward function estimator. 
There exists a universal constant
$c_2>0$, such that
$$
\mathcal{E}(r)\leq c_2\left(\|r-r^*\|^2_{L^2(\mathcal{S}, \ell^2)}\right)^{\frac{1}{3}}.
$$   
\end{corollary}

Comparing Theorem \ref{faster} and Corollary \ref{without_TNC}, we find that the regret bound is improved significantly with the margin-type condition.
It is worth noting that the efficiency gain in our results adjusts automatically with the margin parameter $\alpha$ while remaining independent of the error $\|r-r^*\|^2_{L^2(\mathcal{S}, \ell^2)}$. This improvement is primarily attributed to the use of a high-quality pairwise comparison dataset and is universally applicable to any estimator $r$ employed \citep{tsybakov2007fast, Kim2021Fast}. These findings are consistent with empirical observations in RLHF training \citep{wang2024secrets, chen2024extendingdirect}.
We also obtain the convergence rate of action selection consistency in Section C.2 of the Appendix,
which is a beneficial complement for us to further understand the effects of the margin-type condition.
Based on the margin-type condition, we now present an overview of our main result regarding the regret bound using DNN-based reward estimators.

\begin{theorem}[Informal, Guarantee for Deep Reward Modeling]
\label{thm2_regret_faster_informal}
Consider the deep reward estimator $\hat{r}\in \mathcal{F}_{\operatorname{DNN}}$  where $\mathcal{F}_{\operatorname{DNN}}$ is a class of deep neural networks with width $W=\mathcal{O}(d^{\beta})$ and depth $D=\mathcal{O}(\sqrt{N})$. Then under some regularity assumptions, with probability at least $1-\delta$,
\begin{equation*}
\mathcal{E}(\hat{r})
=
\mathcal{O}
\left(
\left\{ d^{\beta} N^{-\frac{\beta}{d+2\beta}}  + \sqrt{\frac{\log (1/\delta)}{N}} 
\right\}^{\frac{1}{3-2\alpha}}
\right),
\end{equation*}
where $\beta$ is the H\"{o}lder smoothness  parameter for  the reward function $r^*$ and $d$ is the dimension of the input $\mathcal{S}\times \mathcal{A}$.
\end{theorem}
Theorem \ref{thm2_regret_faster_informal} establishes the non-asymptotic regret bound of deep reward estimators in a fully non-parametric setting. By setting the proper depth $D$ and width $W$, the regret of deep reward estimator achieves a rate of $\mathcal{O}(N^{-\beta/[(d+2\beta)(3-2\alpha)]})$. Our analysis provides implications for practitioners on how to choose neural network parameters and construct high-quality comparison datasets to achieve effective reward modeling.

\section{Learning Guarantee of Deep Reward Modeling}\label{Theoretical_Analysis}
{The log-likelihood function for $r$ on the pairwise comparison dataset is written as follows, }
\begin{equation*}
l(r)=
\mathbb{E}\left[\log g\left(y, r(s,a_1)-r(s,a_0)\right)\right].
\end{equation*}
Correspondingly, the empirical log-likelihood is written as,
\begin{equation*}
\label{empirical_l}
\hat{l}(r)=\frac{1}{N}\sum_{i=1} ^N \log g\left(y^i ; r(s^i,a_1^i)-r(s^i,a_0^i)\right).
\end{equation*}
 For a given reward function $r$, the empirical risk $\hat{l}(r)$ is calculated using the observed pairwise comparison data, while the population risk $l(r)$ is the expected value of the risk.
Given the pairwise comparison dataset, we obtain  $\hat{r}$ with the following objectives,
\begin{eqnarray}
\label{eqn:MLE_sec3}
\hat{r} \in \underset{r \in \mathcal{F}_{\operatorname{DNN}}}{\argmax} \ \hat{l}(r).
\end{eqnarray}
To establish the theoretical guarantee of the above estimator, several factors should be considered. 
First, the characteristic of the comparison function $g(y,u)$ captures the relationships between human preference and the underlying reward. 
Second, the smoothness of the true reward function $r^*$ determines how well it can be approximated. 
Most importantly, the configurations of neural networks, \textit{i.e.}, depth and width, need to be leveraged, as they dictate the model's capacity and efficiency to learn complex patterns from finite samples. In the following, we provide definitions and assumptions related to these factors and shape the efficacy of data-driven reward modeling.

\begin{definition}[Comparison Function] 
\label{comparison_function}
A function $g: \Omega \times \mathbb{R} \rightarrow \mathbb{R}^{+}$, where $\Omega$ is a symmetric subset of $\mathbb{R}$ denoting the possible comparison outcomes, is said to be a comparison function if:

(i) For $u \in \mathbb{R}, \int_\Omega g(y, u) d y=1$ if $\Omega$ is continuous, and $\sum_{y \in \Omega} g(y, u)=1$ if $\Omega$ is discrete; 

(ii) $g(y, u)=g(-y, -u), \text{ for any } (y, u) \in \Omega \times \mathbb{R}$; 

(iii) For $y<0, g(y, u)$ is decreasing with respect to $u$, and $g(y, u) \rightarrow 0$ as $u \rightarrow \infty$; 

(iv) $\sup _{u \in \mathbb{R}} g(y, u)<+\infty$, for every $y \in \Omega$;

(v) For every $y$, $\partial^2 \log g(y, u)/\partial u^2 < 0.$
\end{definition}
These conditions guarantee that $g$ functions as a valid probability distribution, exhibiting a symmetric preference structure, stronger preferences for higher relative scores, and log-concavity with respect to $u$. These conditions are mild and widely considered in the literature.  It is straightforward to check that many commonly used models satisfy these conditions, including \textit{BT model}, \textit{Thurstonian model}, \textit{Rao-Kupper model}, and \textit{Davidson model}. 
Next, we describe the characteristics of the reward functions in preference learning.
\begin{assumption}
\label{dynamic_range}
The range of the target reward function $c_{r^*}:=\max_{a \in \mathcal{A}} \sup_{s\in \mathcal{S}} r^*(s,a)$
is finite.
\end{assumption}
\begin{assumption}
\label{Holder_assumption}
(i) The marginal probability measure $\rho_s$ is absolutely continuous with respect to the Lebesgue measure; (ii) For every $a \in \mathcal{A}$, the reward function $r^*(s,a)$ belongs to the 
H\"older class 
$\mathcal{H}^\beta([0,1]^d, c_{\mathcal{H}})$ (See Definition
 \ref{Holder}) for a given smoothness parameter $\beta>0$ and a finite constant $c_{\mathcal{H}}>0$.
\end{assumption}

To ensure the identifiability of  $r^*$, we assume the reward function $\sum_{a\in\mathcal{A}}r^*(s,a) = 0$ for all $s$. This assumption is more of a normalization condition instead of a constraint, as the winning probability is invariant to the shift of reward functions. Similar conditions are considered in \citet{zhu2023principled, rafailov2024direct}. 
In addition, we mention that our theory is general and applies to any underlying reward function, not just normalized ones. For an unknown reward function, we can always transform it into a normalized version, and both the unnormalized and normalized functions lead to the same preference distribution \citep{rafailov2024direct}. Also, we estimate the true reward with the condition $\sum_{a\in\mathcal{A}}\hat{r}(s,a) = 0$ for all $s$. In our theoretical study, the reward estimator is implemented by a fully connected feed-forward neural network consisting of multiple layers of interconnected neurons. Its structure can be described as a composition of linear mappings and activation functions. Specifically, we consider the class of functions $\mathcal{F}_{\operatorname{DNN}}$ consists of $D$-layer feed-forward neural networks that can be expressed as follows,
\begin{equation}
\label{neural}
r(s ,a; \theta)=f_{D+1} \circ f_D \circ \cdots \circ f_2 \circ f_1(s,a),
\end{equation}
where $f_i(x)=\sigma^{(i)} (H^{(i)} x+b^{(i)})$ is the transformation for layer $i$. $H^{(i)}$ and $b^{(i)}$ are the weight matrix and bias vector, respectively. $\sigma^{(i)}$ denotes the ReLU activation function, which is applied to its input elementwisely. We denote the width of the neural network as $W$, which is the maximum of the width of all layers. 
Let $\theta=(H^{(1)}, b^{(1)}, \ldots, H^{(D+1)}, b^{(D+1)})$ represent all the parameters in the neural networks, which consists of $p$ entries.

\subsection{Estimation Within Deep Neural Network Function Class}

With the aforementioned specifications, we start our analysis with the excess risk, which in general stems from two sources: the error from random data realizations and the error from the DNN's limited capacity to represent the target reward. We formalize these intuitions in the following lemma.

\begin{lemma}[Excess risk decomposition]
 The excess risk of $\hat{r}$ is defined and decomposed as 
\label{est_errors}
    \begin{equation*}
        l(r^*)-l(\hat{r})\leq 2 \sup _{r \in \mathcal{F}_{\operatorname{DNN}}}|l(r)-\hat{l}(r)|+\inf _{r \in \mathcal{F}_{\operatorname{DNN}}}\left[l(r^*)-l(r)\right].
    \end{equation*}
\end{lemma}
The first term of the right-hand side is the stochastic error, which measures the difference between the risk $l$ and the empirical counterpart $\hat{l}$ defined over function class $\mathcal{F}_{\operatorname{DNN}}$, evaluating the estimation uncertainty caused by the finite sample size. The second term is the approximation error, which measures how well the function $r^*$ can be approximated using $\mathcal{F}_{\operatorname{DNN}}$ with respect to the likelihood $l(\cdot)$. To assist the following analysis, we define three constants depending on $g(y,u)$ and the range $c_{r^*}$:
$$\kappa_0:=\sup _{y \in \Omega,|u| \leq c_{r^*}}\left|\log g(y , u)\right|, \quad \kappa_1:= \sup _{y \in \Omega,|u| \leq c_{r^*}} \left|\partial\log g(y , u)/\partial u\right|,$$
$$\text{and }\quad \kappa_2:=\inf _{y \in \Omega,|u| \leq c_{r^*}}\left|\partial^2 \log g(y , u)/\partial u^2\right|.$$
These constants are used in the theoretical results, specifying the Lipschitz property and log-concavity invoked from Definition \ref{comparison_function} that ensure the convergence of the maximum likelihood estimator from the deep neural network function class.
\begin{proposition}[Stochastic Error Bound]
\label{stochastic_error}
Under Assumption \ref{dynamic_range}, there exists a universal constant $c_3>0$, with probability at least $1-\delta$,
\begin{equation*}
\begin{aligned}
    &\sup _{r \in \mathcal{F}_{\operatorname{DNN}}}\left|l(r)-\hat{l}(r)\right|\leq\kappa_0\sqrt{\frac{2}{N}}   \left(c_3\sqrt{|\mathcal{A}| Dp \log \left(\frac{ W ((D+1)|\mathcal{A}|N)^{1/D}}{(W!)^{1/{p} }}\right)} \sqrt{\log (1/\delta)}  \right).
\end{aligned}
\end{equation*}
\end{proposition}
If $r^* \in \mathcal{F}_{\operatorname{DNN}}$, 
Proposition \ref{stochastic_error} describes the error of the in-sample learned reward function, measured by the likelihood functional, in comparison to the optimal oracle. This error will scale as $\mathcal{O}(\sqrt{\log (N) / N})$ by considering well-designed network structures from the class $\mathcal{F}_{\operatorname{DNN}}$. It is reasonable that with more collected samples, the DNN can learn the underlying reward function better. Meanwhile, the stochastic error bound increases with the complexity of the function class $\mathcal{F}_{\operatorname{DNN}}$. In other words, once we already know that $r^* \in \mathcal{F}_{\operatorname{DNN}}$ for some network parameters, there is no need to further increase the network's width and depth given the available samples.

\begin{proposition}[Approximation Error Bound]
\label{approx_error}
Let $\mathcal{F}_{\operatorname{DNN}}$ be the deep ReLU neural network class with width and depth specified as
$W=38(\lfloor\beta\rfloor+1)^2 d^{\lfloor\beta\rfloor+1} M_1\left\lceil\log _2(8 M_1)\right\rceil$ and $D=21(\lfloor\beta\rfloor+1)^2 M_2\left\lceil\log _2(8 M_2)\right\rceil$, respectively. Under Assumption \ref{Holder_assumption}, for any $M_1, M_2 \in \mathbb{N}^{+}$, we have
$$\begin{aligned}
    &\inf _{r \in \mathcal{F}_{\operatorname{DNN}}} l(r^*)-l(r)
    \leq
    36\kappa_1 c_{_{\mathcal{H}}}
    (\lfloor\beta\rfloor+1)^2 d^{\lfloor\beta\rfloor + \frac{(\beta \vee 1) }{2}}(M_1 M_2)^{- \frac{2\beta}{d}}.
\end{aligned}$$
\end{proposition}
Proposition \ref{approx_error} demonstrates that the approximation error bound decreases in the size of the function class $\mathcal{F}_{\operatorname{DNN}}$ through two parameters $M_1$ and $M_2$, which are assigned later. This is intuitive since a larger network has greater expressive power. On the other hand, a larger network inflates the stochastic error due to the over-parameterization.

Consequently, it is necessary to carefully design the network structure to strike a balance between the stochastic error and the approximation error.
To do this, we need to appropriately relate $M_1$ and $M_2$ to the sample size $N$ so that an optimal convergence rate of the excess risk bound can be achieved. There are many approaches that result in the same optimal convergence rate while incurring different total numbers of parameters $p$. Note that $p\leq W(d+1)+\left(W^2+W\right)(D-1)+W+1=\mathcal{O}(W^2 D)$, which grows linearly in depth $D$ and quadratically in the width $W$. It is desirable to employ fewer model parameters, and thus, the deep network architectures are preferable to the wide ones. 

Unlike classical regression and classification problems, for pairwise comparison problems, the theoretical results depend on not only $N$ but also $n_{ij}$, the number of comparisons between action $i$ and $j$.
Thus, we consider the graph structure of the dataset, which requires the number of comparisons between each action pair not to be scarce.   
\begin{assumption}[Data Coverage]
\label{assump_datacover}
Let  $\Lambda\in \mathbb{R}^{|\mathcal{A}|\times|\mathcal{A}|}$
be the Laplacian matrix, where $\Lambda_{ij} =- n_{ij}/N$ and $\Lambda_{ii} = \sum_{j \neq i}  n_{ij}/N$.
$\Lambda_{ij}$ is the fraction of sample size in which the pair $(a_i, a_j)$ is compared, while $\Lambda_{ii}$ is the fraction of comparisons  involving $a_i$. 
We assume that there exists a positive constant $ \kappa_{\Lambda}$, such that 
$\lambda_2(\Lambda)>\kappa_{\Lambda}$
where 
$\lambda_2(\Lambda) = \argmin_{w\perp \boldsymbol{1}} w^\top \Lambda w/w^\top w$.
\end{assumption}
If some actions are almost not queried among the data, the spectral gap $1/\lambda_2(\Lambda)$ diverges, and thus the regret of the MLE estimator could not converge \citep[Theorem 3.9]{zhu2023principled}. We refer to Appendix \ref{app_a} for more discussions. Next, we obtain
the optimal bound in terms of the norm $\|\cdot\|^2_{L^2(\mathcal{S}, \ell^2)}$, which is defined in (\ref{L2_distance}).

\begin{theorem}[Non-asymptotic Estimation Error Bound]
\label{non_asymp_error}
Given the network parameters specified  in Proposition \ref{approx_error}, with $M_1=1$ and $M_2=\left\lfloor N^{d/(2d+4\beta)}\right\rfloor$, 
under Assumptions \ref{dynamic_range}, \ref{Holder_assumption} and \ref{assump_datacover}, there exists a universal constant $c_4>0$, with probability at least $1-\delta$,
\begin{equation*}
\begin{aligned}
    \|\hat{r}-r^*\|^2_{L^2(\mathcal{S}, \ell^2)} \leq 2\sqrt{2}\frac{\kappa_0}{\kappa_2 \kappa_{\Lambda}}  \bigg(c_4\sqrt{ |\mathcal{A}|}(\lfloor\beta\rfloor+1)^4 d^{\lfloor\beta\rfloor+1}
    (\log (N))^2 N^{-\frac{\beta}{d+2\beta}}  +   \sqrt{\frac{\log (1/\delta)}{N}}  \bigg).
\end{aligned}
\end{equation*}
\end{theorem}
Theorem \ref{non_asymp_error} presents the non-asymptotic convergence rate of $\|\hat{r}-r^*\|^2_{L^2(\mathcal{S}, \ell^2)}$. 
Unlike deep regression and classification problems, the convergence of excess risk does not directly guarantee the functional convergence of the estimated reward unless additional constraints on the dataset structure are imposed (see Assumption \ref{assump_datacover}). This condition is also considered by \citet{zhu2023principled}.
Combining Theorems \ref{faster} and \ref{non_asymp_error}, we obtain the regret bound for the decision maker $\pi_{\hat{r}}(s)$ induced by the deep reward estimator under the margin-type condition.

\begin{theorem}
\label{regret_faster}
Let $\mathcal{F}_{\operatorname{DNN}}$ be the deep ReLU neural networks class with width and depth, respectively, specified as
$$ W=114(\lfloor\beta\rfloor+1)^2 d^{\lfloor\beta\rfloor+1}, \quad \text{and} \quad
D=21(\lfloor\beta\rfloor+1)^2 N^{\frac{d}{2 d+4 \beta}}\lceil\log _2(8 N^{\frac{d}{2 d+4 \beta}})\rceil.$$ Then given the MLE estimator $\hat{r} \in \mathcal{F}_{\operatorname{DNN}}$, under Assumptions 
\ref{noisep},
\ref{dynamic_range}, \ref{Holder_assumption} and \ref{assump_datacover},, there exists a universal constant $c_5$, such that with probability at least $1-\delta$,
\begin{equation*}
\begin{aligned}
    \mathcal{E}(\hat{r})\leq c_5\left(\frac{\kappa_0\sqrt{ |\mathcal{A}|}(\lfloor\beta\rfloor+1)^4 d^{\lfloor\beta\rfloor+1}(\log N)^2}{\kappa_2\kappa_{\Lambda}}\right)^{\frac{1}{3-2\alpha}} 
     N^{-\frac{\beta}{(d+2\beta)(3-2\alpha)}}  +   \left(\frac{\kappa^2_0\log \left(\frac{1}{\delta}\right)}{\kappa_2^2 \kappa^2_{\Lambda} N}\right)^{\frac{1}{2(3-2\alpha)}}.
\end{aligned}
\end{equation*}
\end{theorem}
Our results reveal that deep neural network reward estimators offer satisfactory solutions with explicit theoretical guarantees in this general setting.
To achieve the fast convergence rate, it is essential to train the model with sufficient data and select an appropriate network structure, adhering to the guidelines for width and depth selection provided in Theorem \ref{regret_faster}. To be specific, the width is a multiple of $d^{\lfloor\beta\rfloor+1}$, a polynomial of the feature dimension $d$; The depth is proportional to $\sqrt{N}$, as $d \gg \beta$. 
To the best of our knowledge, we are the first to study the non-parametric analysis of deep reward modeling, providing more insights into the DNN-based reward modeling in RLHF.

\section{Experiments}\label{sec:exp}
In this section, we conduct two experiments to provide the numerical evidence to support our theoretical results. The comparison dataset is denoted as  $\mathcal{D}_N  = \{s^i,a^i_1,a^i_0,y^i\}_{i=1}^N$. The state $s$ is sampled independently from a uniform distribution over $[0,1]^d$ with $d=10$. The action pair $a^i_1, a^i_0 \in \mathcal{A}$ where $\mathcal{A} = \{0, 1\}$.
The true reward functions are specified as $r^*(s,a_1) = 2\sin(4  \phi(s)^\top w^*)$ and $r^*(s,a_0) = - r^*(s,a_1)$. The true weight $w^*$ is generated randomly from a standard normal distribution. $\phi(s) = (\sin(s_1), \ldots, \sin(s_d))^\top$ is a non-linear transformation. It is worth noting that the identification condition, $r^*(s,a_1) + r^*(s,a_0) = 0$ for all $s$, is satisfied.
We consider both BT and Thurstonian models, with the preference $y^i$ generated as described in Examples \ref{BT} and \ref{thurstonian}, respectively. Additional experiments on various reward functions and details are presented in Appendix \ref{appendix_experiment}.

\subsection{Experiment 1: Neural Network Configurations}\label{sec:4.1}

We investigate how network architecture balances approximation and stochastic errors in reward modeling. We test rectangular networks with widths ranging from $2^2$ to $2^{12}$ and depths from 3 to 13, training each configuration on $(N_{train},N_{eval},N_{test})=(2^{14},2^{13},2^{14})$ with 50 replications. resulting in parameter counts from $64$ to about $2\times 10^7$. This wide range of architectures enables us to examine the trade-off between approximation and stochastic errors described in Theorem \ref{non_asymp_error}.

\begin{figure}[h!]
    \centering
    \includegraphics[width=0.8\textwidth]{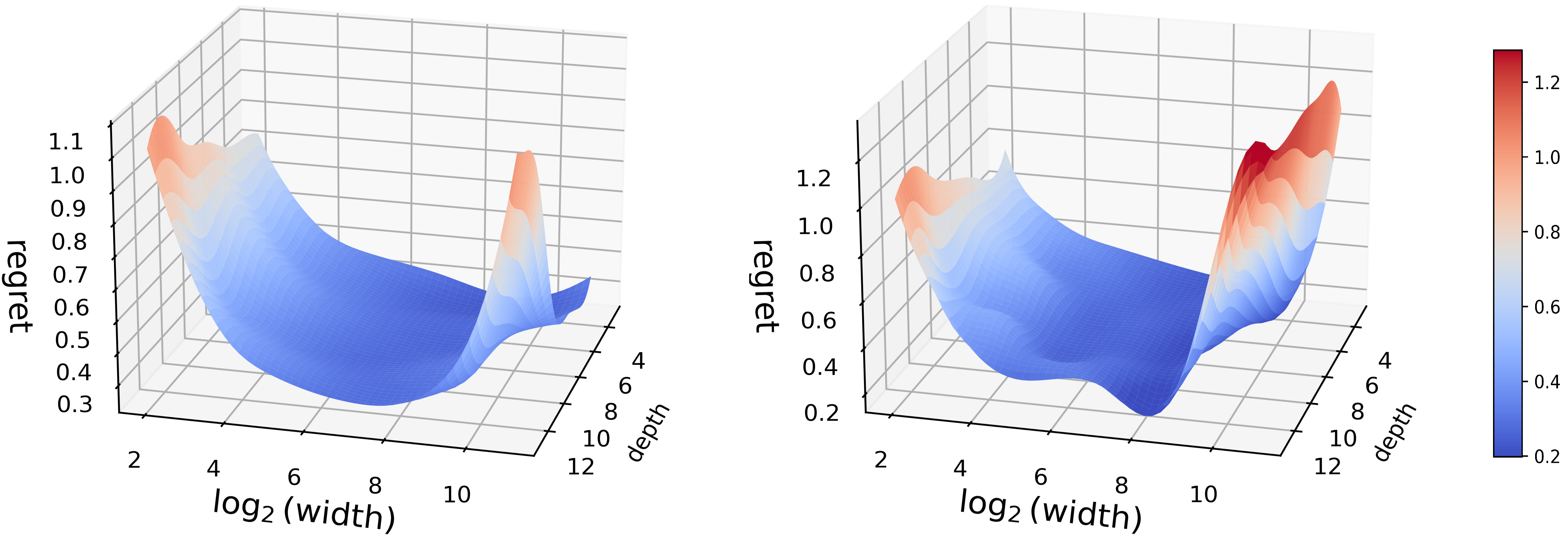}
    \caption{Surface plot of regret for both BT (upper panel) and Thurstonian (lower panel) models for sinusoidal reward functions across different neural network configurations.}
    \label{fig:result1}
\end{figure}

Figure \ref{fig:result1} shows that regret initially decreases with deeper and wider networks, as the approximation capability of networks increases. However, the performance degrades beyond some near-optimal configurations, as stochastic error becomes dominant. Notably, a broad range of architectures in a relatively flat region of the parameter space achieves near-minimal regret, demonstrating DNN's adaptability to unknown function complexity. This mitigates the need for precise knowledge of the smoothness to achieve strong empirical performance \citep{jiao2023deep, lee2019wide, havrilla2024understanding}.

\subsection{Experiment 2: High-quality Pairwise Comparison Dataset}
We examine how label ambiguity affects reward modeling by systematically corrupting pairwise comparison data. Using the same dataset sizes as in Section \ref{sec:4.1}, reflecting cases where preference data contain inconsistent or ambiguous labels. For each noise level $m \in \{0.1, 0.2, \dots, 0.7, 0.8\}$, $m\times 100\%$ samples from the training and evaluation datasets are selected, with the value of their conditional probabilities $\mathbb{P}(y>0 \mid s,a_1,a_0)$ are now drawn uniformly from $[0.4, 0.6]$. This creates a spectrum of data quality from clear preferences $(m=0)$ to heavily corrupted as $m$ grows, as visualized in Figure \ref{fig:conditioal_probabilities} of the Appendix. We fix the depth as $4$, and the width as $64$, and the same training protocol from Experiment 1, evaluating performance on an uncontaminated test set to isolate the effect of label noise.

\begin{figure}[h!]
    \centering
    \includegraphics[width=0.8\textwidth]{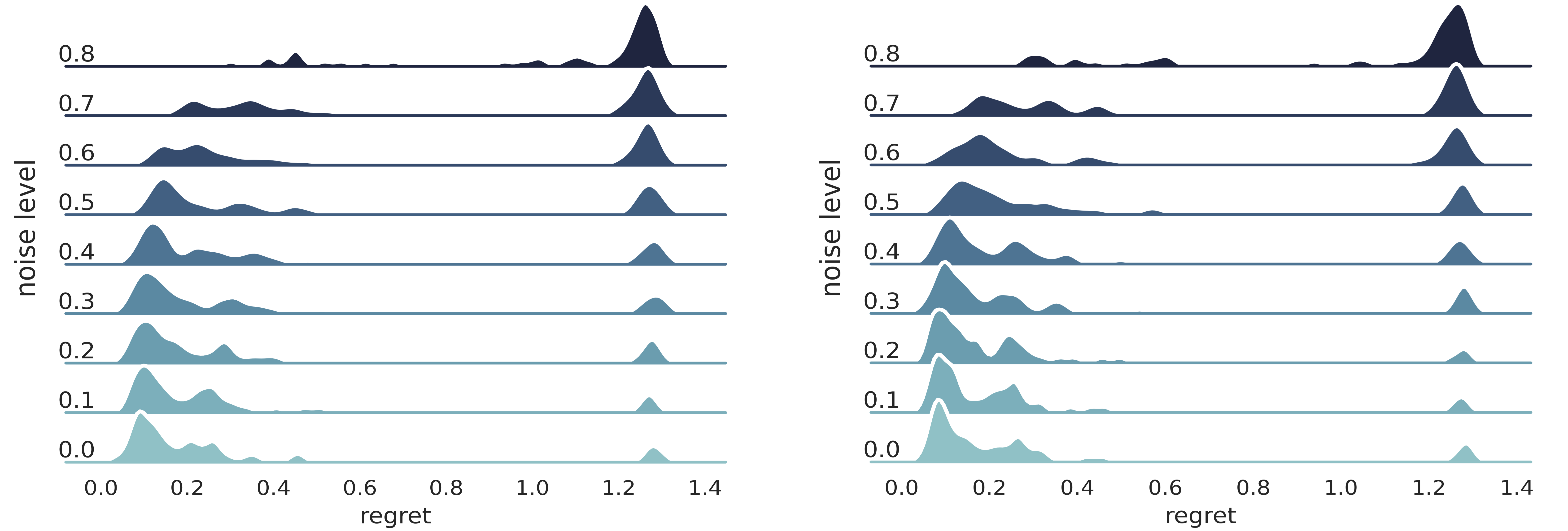}
    \caption{Empirical distribution of regret for both BT (upper panel) and Thurstonian (lower panel) models for sinusoidal reward functions under different noise levels. A larger noise level implies a lower quality of the comparison dataset, resulting in higher regret.}
    \label{fig:result2}
\end{figure}

As shown in Figure \ref{fig:result2}, lower noise levels, which indicate stronger and more consistent human preferences, yield significantly lower regret and more concentrated distributions. In contrast, higher noise levels, typically associated with ambiguous or inconsistent preferences, result in higher regret and broader distributions.
These findings align with recent studies suggesting that focusing on clear, well-defined preference pairs leads to more reliable model performance \citep{ouyang2022training,wang2024secrets}.
The results empirically validate the importance of careful data curation and the potential necessity of filtering out ambiguous samples in practical applications, particularly when dealing with subjective human judgments in preference learning tasks.

\section{Related Works}\label{sec:related_work}
\paragraph*{Reinforcement Learning from Human Feedback} 

Recent theoretical advances in offline RLHF include the development of reward-based preference models by \citet{zhu2023principled} for linear models and \citet{zhan2023provable} for general function classes, though non-parametric analysis of DNN-based reward estimators remains understudied. While these works establish foundational frameworks, the impact of human belief quality has only recently emerged as a crucial consideration. \citet{wang2024secrets} proposed preference label correction protocols, and \citet{zhong2024provable} addressed preference heterogeneity through meta-learning approaches, yet a comprehensive theoretical framework for quantifying human belief in RLHF remains to be developed.

\paragraph*{Margin Condition for Human Preference}
The efficiency of RLHF is largely attributed to strong human preferences reflected in pairwise comparison data \citep{zhong2024provable}.
This concept parallels established noise conditions in classification theory, particularly Massart and Tsybakov noise conditions, which bound excess misclassification error \citep{tsybakov2004optimal, diakonikolas2022learning}. 
Similar  conditions have proven valuable across various domains, from individualized treatment analysis \citep{qian2011performance} to 
linear bandit problems \citep{goldenshluger2013linear, bastani2020online}. 
Recent empirical studies aim to enhance data quality and address the problem of ambiguous samples in practice
\citep{wang2024secrets,liu2024dual,zhan2023how}.
Our study provides a theoretical foundation for this phenomenon by introducing margin-type conditions to reward function learning.
\paragraph*{Convergence Analysis for Deep Neural Network Estimators} 
Recent studies have proposed theoretical results on the generalization performance of DNNs, which include norm-based generalization bounds \citep{bartlett2017spectrally, hieber2020aos} and
algorithm-based generalization bounds \citep{wang2022generalization}.
The optimization aspects are beyond the scope of this study, and we refer to \citet{allen2019convergence,lyu2021gradient} for more discussion.

\section{Conclusion}\label{sec:conclusion}
We establish non-asymptotic regret bounds for DNN-based reward estimation, showing that properly configured architectures balance stochastic and approximation errors for optimal convergence.
This fully non-parametric approach addresses the issue of reward model mis-specification in linear reward settings while remaining applicable to various pairwise comparison models. Furthermore, we derive a sharper bound based on a margin condition related to the comparison dataset, rather than making direct assumptions about the unknown reward function. This margin condition is practical and verifiable through estimated winning probabilities, whereas assumptions regarding the unknown reward function are often unverifiable in practice. Our theoretical framework emphasizes the importance of controlling the effects of ambiguous data, aligning with insights from prior research on learning halfspaces and active learning \citep{yan2017revisiting,zhang2020efficient}. Looking ahead, it would be valuable to extend our results to trajectory-based comparisons within Markov decision process frameworks \citep{zhu2023principled}. These considerations could provide further insights when applied to broader reinforcement learning tasks.

\bibliography{reference}
\bibliographystyle{apalike}

\newpage
\appendix

{\Large
\textbf{Appendix}
}

In this Appendix, we present the technical details of the proof of theorems and lemmas and provide supporting definitions.
\section{Technical Notations}\label{app_a}
\paragraph*{Notations:}
For sequences $\left\{b_n\right\}_{n \in \mathbb{N}}$ and $\left\{c_n\right\}_{n \in \mathbb{N}}$, we say $b_n = \mathcal{O}(c_n)$ if there exists an absolute constant $k$ and $n_0 \in \mathbb{N}$ such that $b_n \leq k c_n$ for all $n > n_0$. And we say $b_n = \Theta(c_n)$ if there exists absolute constants $k_1,k_2$ and $n_0 \in \mathbb{N}$ such that $ k_1 c_n\leq b_n \leq k_2 c_n$ for all $n > n_0$. Let $\lceil u\rceil$ denote the smallest integer that is no less than $u$, and $\lfloor u \rfloor$ denote the greatest integer that is no greater than $u$.
For a Lebesgue measurable subset $\mathcal{S} \subseteq \mathbb{R}^d$, by $L^q\left(\mathcal{S}, \ell^p\right)$ we denote the function norm for all real-valued functions $f: \mathcal{S}\times\mathcal{A} \rightarrow \mathbb{R}$, such that for every $a\in \mathcal{A}$, $f(\cdot,a)$ is Lebesgue measurable on $\mathcal{S}$ and
\begin{equation}
\label{L2_distance}
\|f\|_{L^q(\mathcal{S}, \ell^p)}:= \begin{cases}\left(\displaystyle \int_{\mathcal{S}} \left(\sum_{a\in \mathcal{A}}|f(s,a)|^p\right)^\frac{q}{p} d \rho_s\right)^{\frac{1}{q}}, & 1 \leq q<+\infty, \\ \operatorname{ess} \sup _{s \in \mathcal{S}} \left(\sum_{a\in \mathcal{A}}|f(s,a)|^p\right)^\frac{1}{p}, & q=+\infty\end{cases}
\end{equation}
is finite.

\begin{definition}[H\"{o}lder Function Class]
\label{Holder}
For  $\beta, c_{\mathcal{H}}>0$, and a domain $\mathcal{X} \in \mathbb{R}^d$, the H\"older function class $\mathcal{H}^\beta(\mathcal{X}, c_{\mathcal{H}})$ is defined by
$$
\begin{aligned}
    \mathcal{H}^\beta(\mathcal{X}, c_{\mathcal{H}})&=\bigg \{f: \mathcal{X} \rightarrow \mathbb{R}, \max _{\|\boldsymbol{\omega}\|_1 \leq \lfloor\beta\rfloor}\|\partial^{\boldsymbol{\omega}} f\|_{\infty} \leq c_{\mathcal{H}}, \max _{\|\boldsymbol{\omega}\|_1=\lfloor\beta\rfloor} \sup _{x \neq x^\prime} \frac{|\partial^{\boldsymbol{\omega}} f(x)-\partial^{\boldsymbol{\omega}} f(x^\prime)|}{\|x-x^\prime\|_2^{\beta-\lfloor\beta\rfloor }} \leq c_{\mathcal{H}} \bigg\},
\end{aligned}
$$
where $\boldsymbol{\omega}=(\omega_1, \ldots, \omega_d)^{\top}$ is a vector of non-negative integers, $\|\boldsymbol{\omega}\|_1:=\sum_{i=1}^d \omega_i$, and $\partial^{\boldsymbol{\omega}}=\partial^{\omega_1} \cdots \partial^{\omega_d}$ denotes the partial derivative operator.
\end{definition}

\begin{definition}[Covering Number of a Function Class]
Let $\mathcal{F}$ be a class of functions $: \mathcal{X} \rightarrow \mathbb{R}$. For a given $\epsilon>0$, we denote $\mathcal{N}\left(\mathcal{F}, \epsilon,\|\cdot\|\right)$ as the covering number of $\mathcal{F}$ with radius $\delta$ under some norm $\|\cdot\|$ as the least cardinality of a subset $\mathcal{F}^\prime \subseteq \mathcal{F}$, satisfying
$$
\sup _{f \in \mathcal{F}} \min _{f^\prime \in \mathcal{F}^\prime}\|f-f^\prime\| \leq \delta.
$$
This quantity measures the minimum number of functions in $\mathcal{F}$ needed to cover the set of functions within a distance of $\delta$ under the norm $\|\cdot\|$.
\end{definition}

\paragraph*{Laplacian Matrix and Graph Structure:}
The connectivity of the comparison graph plays a crucial role in estimating from pairwise data \citep{mohar1991laplacian}, which relates to the second smallest eigenvalue of $\Lambda$, we denote it as $\lambda_2(\Lambda):=\min_{w\perp \boldsymbol{1}} w^\top \Lambda w/w^\top w$. The pair of actions being compared needs to be carefully selected for effective reward modeling, ensuring that $\lambda_2(\Lambda)$ is not excessively small.
If some actions are almost not queried among the data, the spectral gap $1/\lambda_2(\Lambda)$ diverges and thus makes the MLE estimator degenerate in terms of regret. 
This happens if the number of comparisons between some pairs is too scarce, as shown in Example \ref{lambda_blows_up}, where $1/\lambda_2(L)$ can blow up to an order of $\Theta(n)$.
\begin{example}
\label{lambda_blows_up}
Suppose there are four actions in $\mathcal{A}$. Let $n=2t+1$, we query $(a_1,a_2), (a_2,a_3)$ for $t$ times each, and $(a_3,a_4)$ only once. The Laplacian matrix of this pairwise comparison design is 
   \begin{equation*}
\Lambda = \frac{1}{n}\left[\begin{array}{cccc}
t & -t & 0 & 0 \\
-t & 2 t & -t & 0 \\
0 & -t & t+1 & -1 \\
0 & 0 & -1 & 1
\end{array}\right]
\end{equation*}
It is clear that $1/\lambda_2(\Lambda) \asymp 3(2t+1)/2 = \Theta(n)$, which blows up the error $\|\hat{r}-r^*\|^2_{L^2(\mathcal{S}, \ell^2)}$, although the excess risk is still under control. 
\end{example}
It is worth pointing out that the optimal choice of $\Lambda$ should satisfy that $\lambda_2(\Lambda)=\Theta(1/|\mathcal{A}|)$. As $\Lambda$ is the Laplacian matrix whose  $\operatorname{trace}(\Lambda)=2$,  $\lambda_2(\Lambda) \leq 2/(|\mathcal{A}|-1)$. A natural choice to reach this bound is to require every pair of actions to be compared equally. In the terminology of the graph, this choice is referred to as the \textit{complete graph}. In addition to the \textit{complete graph}, there are several graphs  satisfying $\lambda_2(\Lambda)=\Theta(1/|\mathcal{A}|)$,  such as the \textit{complete bipartite} graph and \textit{star} graph. In contrast, the \textit{path} graph and \textit{cycle} graph lead to $\lambda_2(\Lambda)=\Theta(1/|\mathcal{A}|^3)$ \citep{shah2016estimation}. Therefore, effective reward modeling requires careful selection of the pairwise comparison subset.

\section{Proof of Lemma \ref{noiser}}
We define $$G(u):= \int_0^\infty g(y , u)dy-\frac{1}{2}.$$
By definition of the comparison function $g$, it is straightforward that $G(0) \leq 0$, so we can always find a tangent line to $G(u)$ crossing the origin on $\mathbb{R}^+$, \textit{i.e.,} there exists a constant $\kappa \leq \sup _{|u| \leq c_{r^*}}|\frac{d}{d u}G(u)|$, such that $G(u)\leq \kappa u$. Note that $G(u/\kappa)\leq u$ also holds for all $u\in \mathbb{R}^+$. Then the set\begin{equation*}
    \left\{s\in \mathcal{S}:G(r^*(s,\pi_{r^*}(s))-r^*(s,a'))\leq G\left(\frac{t}{\kappa}\right)\right\} \subseteq \left\{s\in \mathcal{S}:G(r^*(s,\pi_{r^*}(s))-r^*(s,a'))\leq t\right\}.
\end{equation*}Also, due to the monotonicity of the function $G(u)$, the set 
\begin{equation*}
        \left\{s\in \mathcal{S}:G(r^*(s,\pi_{r^*}(s))-r^*(s,a')) \leq G\left(\frac{t}{\kappa}\right)\right\}=\left\{s\in \mathcal{S}: r^*(s,\pi_{r^*}(s))-r^*(s,a') \leq \frac{t}{\kappa}\right\}.
\end{equation*}
Thus, 
$$ 
\int_{\mathcal{S}} \1\left\{r^*(s,\pi_{r^*}(s))-r^*(s,a') \leq \frac{t}{\kappa}\right\} d \rho_{s} \leq c_g t^{\frac{\alpha}{1-\alpha}}.
$$  
Replace  $t/\kappa$ with $t$ and update the constant results in the desired inequality.
\begin{example}
Here we consider the \textit{BT model} as an example (see Example \ref{BT}). We deliberately set $t'=2t$, where $t$ is the probability gap in Assumption \ref{noisep}.

\begin{equation*}
\begin{aligned}
     &\int_{\mathcal{S}} \1\left\{r^*(s,r^*(s,\pi_{r^*}(s)))-r^*(s,a') \leq t^\prime\right\} d \rho_{s} \\
    & = \int_{\mathcal{S}} \1 \left\{\frac{\exp(r^*(s,\pi_{r^*}(s)))}{\exp(r^*(s,\pi_{r^*}(s)))+\exp(r^(s,a'))}-\frac{1}{2}\leq\frac{1}{2}\frac{e^{t^\prime}-1}{e^{t^\prime}+1}\right\} d\rho_s. \\
    &= \mathbb{P}_{\mathcal{S}}\left(\mathbb{P}(y>0 \mid s,a_1=\pi_{r^*}(s),a_0=a^\prime)-1/2\leq \frac{1}{2}\frac{e^{t^\prime}-1}{e^{t^\prime}+1}\right)\\
    &\leq \mathbb{P}_{\mathcal{S}}\left(\mathbb{P}(y>0 \mid s,a_1=\pi_{r^*}(s),a_0=a^\prime)-1/2\leq t^\prime/4\right)\\
    &\leq (1/4)^{\frac{\alpha}{1-\alpha}}c_g {t^\prime}^{\frac{\alpha}{1-\alpha}}.
\end{aligned}
\end{equation*}
The first step holds since $\frac{1}{2}\frac{e^{t^\prime}-1}{e^{t^\prime}+1}$ is monotonically increasing. The third is due to $\frac{t^\prime}{4} \geq \frac{1}{2}\frac{e^{t^\prime}-1}{e^{t^\prime}+1}$ for all $t'\in (0,1)$. Replacing the notation $t'$ with $t$ yields the desired inequality.
\end{example}

\begin{example}
In the \textit{Thurstonian model} (see Example \ref{thurstonian}), we set $t'=\sqrt{\pi/2} \ t$, where $t$ is the probability gap in Assumption \ref{noisep}.
\begin{equation*}
\begin{aligned}
     &\int_{\mathcal{S}} \1\left\{r^*(s,r^*(s,\pi_{r^*}(s)))-r^*(s,a') \leq t^\prime\right\} d \rho_{s} \\
    &= \int_{\mathcal{S}} \1 \left\{\frac{\exp(-(r^*(s,r^*(s,\pi_{r^*}(s)))-r^*(s,a'))^2 / 2)}{\sqrt{2 \pi}}-\frac{1}{2}\leq \frac{e^{-{t^\prime}^2 / 2}}{\sqrt{2 \pi}} -\frac{1}{2}\right\} d\rho_s. \\
    &= \mathbb{P}_{\mathcal{S}}\left(\mathbb{P}(y>0 \mid s,a_1=\pi_{r^*}(s),a_0=a^\prime)-1/2\leq \frac{e^{-{t^\prime}^2 / 2}}{\sqrt{2 \pi}}-\frac{1}{2}\right)\\
    &\leq \mathbb{P}_{\mathcal{S}}\left(\mathbb{P}(y>0 \mid s,a_1=\pi_{r^*}(s),a_0=a^\prime)-1/2\leq t^\prime/\sqrt{2 \pi}\right)\\
    &\leq (1/\sqrt{2 \pi})^{\frac{\alpha}{1-\alpha}}c_g {t^\prime}^{\frac{\alpha}{1-\alpha}}.
\end{aligned}
\end{equation*}
The first step holds since $\frac{e^{-{t^\prime}^2 / 2}}{\sqrt{2 \pi}} -\frac{1}{2}$ is monotonically increasing. The third is due to $t^\prime/\sqrt{2 \pi} \geq \frac{e^{-{t^\prime}^2 / 2}}{\sqrt{2 \pi}} -\frac{1}{2}$ for all $t'\in (0,1)$. Replacing the notation $t'$ with $t$ yields the desired inequality.
\end{example}

The constant $c_g>0$ in Assumption 2.1 is determined by the data distribution and the specifications of the comparison model. It is important to note that  $c_g$ is finite and does not affect the rate of the preceding theoretical bounds.  To illustrate this, we calculate the exact value of $c_g$ using the following example.
\begin{example}
In a \textit{BT} model with two actions, let $a^\prime = \mathcal{A} \backslash \pi_{r^*}(s)$. Suppose that there exists a distribution $\rho_s$ such that the random variable $Z = 1/(1+\exp{(r^*(s, \pi_{r^*}(s))-r^*(s, a^{\prime}))})$ is uniformly distributed on $(1/2,1)$. Let the noise exponent $\alpha =1/2$, then Assumption \ref{noisep} implies
$$\int_{\mathcal{S}} \1\left\{Z - \frac{1}{2} \leq t \right\} d \rho_{s} \leq c_g t.$$
Since $Z-\frac{1}{2}$ is uniformly distributed on $(0,1/2)$, \textit{i.e.}, $\mathbb{P}(0<Z-\frac{1}{2}<t) = 2t$ for $0 < t < 1/2$, we immediately know $c_g=2$.
\end{example}

\section{Proof of Theorem \ref{faster}}
By the definition of regret,
\begin{equation*}
\begin{aligned}
\mathcal{E}(r)=& \int_{\mathcal{S}} \left( r^*(s,\pi_{r^*}(s)) - r^*(s,\pi_{r^*}(s))   \right) d \rho_{s}\\
= & \int_{\pi_{r^*}(s)\neq \pi_{\hat{r}}(s)} \left( r^*(s, \pi_{r^*}(s)) - r^*(s, \pi_{\hat{r}}(s))   \right) d \rho_{s}.
\end{aligned}
\end{equation*}
Now, we define two sets, given any $\eta \in (0,1)$,
\begin{equation*}
\begin{aligned}
\mathcal{S}_1=& \{s\in \mathcal{S}: 0< r^*(s, \pi_{r^*}(s)) - \max_{a\in \mathcal{A}/\pi_{r^*}(s)}r^*_a(s)   \leq \eta\};\\
\mathcal{S}_2=& \{s\in \mathcal{S}: r^*(s, \pi_{r^*}(s)) - \max_{a\in \mathcal{A}/\pi_{r^*}(s)}r^*_a(s)   > \eta\}. \\
\end{aligned}
\end{equation*}
It is worth noting that the two sets $\mathcal{S}_1$ and $\mathcal{S}_2$ are the complement of each other, and they are independent of any reward estimator $r$. Then, we can decompose the performance loss as follows,
{\small\begin{equation}
\label{decomposition}
\begin{aligned}
\mathcal{E}(r)=& \int_{\pi_{r^*}(s)\neq \pi_{r}(s) \cap \mathcal{S}_1} \left( r^*(s, \pi_{r^*}(s)) - r^*(s, \pi_{r}(s))   \right) d \rho_{s}  + \int_{\pi_{r^*}(s)\neq \pi_{r}(s) \cap \mathcal{S}_2} \left( r^*(s, \pi_{r^*}(s)) - r^*(s, \pi_{r}(s))   \right) d \rho_{s}\\
=&\int_{\pi_{r^*}(s)\neq \pi_{r}(s) \cap \mathcal{S}_1} \left( r^*(s, \pi_{r^*}(s)) - r^*(s, \pi_{r}(s))   \right) d \rho_{s} \\
&\quad\quad\quad\quad+ \int_{\pi_{r^*}(s)\neq \pi_{r}(s) \cap \mathcal{S}_2} \left( r^*(s, \pi_{r^*}(s)) - r^*(s, \pi_{r}(s))   \right)  \1\left\{ \sum_{a\in \mathcal{A}}\left|r(s, a) - r^*(s, a)\right| \leq \eta   \right\} d \rho_{s}\\
&\quad\quad\quad\quad+ \int_{\pi_{r^*}(s)\neq \pi_{r}(s) \cap \mathcal{S}_2} \left( r^*(s, \pi_{r^*}(s)) - r^*(s, \pi_{r}(s))   \right)  \1\left\{ \sum_{a\in \mathcal{A}}\left|r(s, a) - r^*(s, a)\right| \geq \eta   \right\} d \rho_{s}.\\
\end{aligned}
\end{equation}}
Note that the second term in the last step is $0$ since there is no regret loss as long as the estimated action is the optimal action.
\begin{equation*}
\begin{aligned}
&\mathcal{E}(r) \leq  \eta \int_{\mathcal{S}} \1\left\{0<r^*(s, \pi_{r^*}(s))-\max_{a\in \mathcal{A}/\pi_{r^*}(s)}r^*_a(s) \leq \eta \right\} d \rho_{s}+0\\
&\quad\quad\quad\quad\quad\quad\quad\quad\quad\quad\quad\quad\quad\quad +c_{r^*}\int_{\pi_{r^*}(s)\neq \pi_{r}(s) \cap \mathcal{S}_2}\1\left\{ \sum_{a\in \mathcal{A}}\left|r(s, a) - r^*(s, a)\right| \geq \eta \right\} d \rho_{s}.
\end{aligned}
\end{equation*}
Recall that with Lemma \ref{noiser}, for a non-negative random variable $X$, and non-decreasing function $\varphi(u)>0$, the Markov inequality states
$\varphi(u)\mathbb{P}(X \geq u) \leq \mathbb{E}(\varphi(X))$. Let $\varphi(\eta)= \eta^{2},$ where $\alpha \in (0,1)$. Then, 
{\small
\begin{equation*}
\begin{aligned}
    \eta^{2} \int_{\pi_{r^*}(s)\neq \pi_{r}(s) \cap \mathcal{S}_2}    \1\left\{ \sum_{a\in \mathcal{A}}\left|r(s, a) - r^*(s, a)\right| \geq \eta   \right\} d \rho_{s} 
    &\leq \eta^{2}\int_{\mathcal{S}} \1\left\{ \sum_{a\in \mathcal{A}}\left|r(s, a) - r^*(s, a)\right| \geq \eta   \right\} d \rho_{s} \\
    &\leq  \int_{\mathcal{S}} \left(\sum_{a\in \mathcal{A}}\left|r(s, a) - r^*(s, a)\right|\right)^2 d \rho_{s},
\end{aligned}
\end{equation*}}
where we apply the Markov inequality in the last step, taking an expectation over the state space $\mathcal{S}$. Note that $\|r-r^*\|^2_{L^2(\mathcal{S}, \ell^1)}\leq |\mathcal{A}| \|r-r^*\|^2_{L^2(\mathcal{S}, \ell^2)}$, then
\begin{equation*}
\mathcal{E}(r) \leq \eta \cdot c_g \eta^{\frac{\alpha}{1-\alpha}}+ (\frac{1}{\eta})^{2}|\mathcal{A}| c_{r^*} \|r-r^*\|^2_{L^2(\mathcal{S}, \ell^2)}.
\end{equation*}
To balance the two terms above, 
we choose $$\eta = \left(\frac{|\mathcal{A}| c_{r^*}}{c_g}\|r-r^*\|^2_{L^2(\mathcal{S}, \ell^2)}\right)^{\frac{1-\alpha}{3-2\alpha}}.$$
Consequently, we have
\begin{equation*}
\begin{aligned}
   \mathcal{E}(r)\leq &  c_1 \left(\|r-r^*\|^2_{L^2(\mathcal{S}, \ell^2)}\right)^{\frac{1}{3-2\alpha}}
\end{aligned}
\end{equation*}
with $c_1 = \left(\frac{|\mathcal{A}| c_{r^*}}{c_g}\right)^{\frac{1-\alpha}{3-2\alpha}}$.
$\square$

\subsection{Proof of Corollary \ref{without_TNC}}

The proof generally follows the proof of Theorem \ref{without_TNC} but without any margin condition on the state distribution. Starting from the second step of (\ref{decomposition}), we have
{\small
\begin{equation*}
\begin{aligned}
\mathcal{E}(r)=& \int_{\pi_{r^*}(s)\neq \pi_{r}(s) \cap \mathcal{S}_1} \left( r^*(s, \pi_{r^*}(s)) - r^*(s, \pi_{r}(s))   \right) d \rho_{s} + \int_{\pi_{r^*}(s)\neq \pi_{r}(s) \cap \mathcal{S}_2} \left( r^*(s, \pi_{r^*}(s)) - r^*(s, \pi_{r}(s))   \right) d \rho_{s}\\
=&\int_{\pi_{r^*}(s)\neq \pi_{r}(s) \cap \mathcal{S}_1} \left( r^*(s, \pi_{r^*}(s)) - r^*(s, \pi_{r}(s))   \right) d \rho_{s} \\
&\quad\quad\quad\quad+ \int_{\pi_{r^*}(s)\neq \pi_{r}(s) \cap \mathcal{S}_2} \left( r^*(s, \pi_{r^*}(s)) - r^*(s, \pi_{r}(s))   \right)  \1\left\{ \sum_{a\in \mathcal{A}}\left|r(s, a) - r^*(s, a)\right| \leq \eta   \right\} d \rho_{s}\\
&\quad\quad\quad\quad+ \int_{\pi_{r^*}(s)\neq \pi_{r}(s) \cap \mathcal{S}_2} \left( r^*(s, \pi_{r^*}(s)) - r^*(s, \pi_{r}(s))   \right) \1\left\{ \sum_{a\in \mathcal{A}}\left|r(s, a) - r^*(s, a)\right| \geq \eta  \right\}d \rho_{s}\\
\leq & \eta+0+c_{r^*}\int_{\pi_{r^*}(s)\neq \pi_{r}(s) \cap \mathcal{S}_2} \1\left\{ \sum_{a\in \mathcal{A}}\left|r(s, a) - r^*(s, a)\right| \geq \eta   \right\}d \rho_{s}\\
\leq & \eta+  (\frac{1}{\eta})^{2} |\mathcal{A}|c_{r^*} \|r-r^*\|^2_{L^2(\mathcal{S}, \ell^1)}.
\end{aligned}
\end{equation*}}
Note that $\|r-r^*\|^2_{L^2(\mathcal{S}, \ell^1)}\leq |\mathcal{A}| \|r-r^*\|^2_{L^2(\mathcal{S}, \ell^2)}$. Then
with $\eta = (|\mathcal{A}|c_{r^*}\|r-r^*\|^2_{L^2(\mathcal{S}, \ell^2)})^{1/3}$ and  $c_2= \left(|\mathcal{A}|c_{r^*}\right)^{1/3}$,
we have
\begin{equation*}
\mathcal{E}(C_{r}) \leq \eta+  (\frac{1}{\eta})^{2}|\mathcal{A}|c_{r^*} \|r-r^*\|^2_{L^2(\mathcal{S}, \ell^2)}\leq  c_2\left(\|r-r^*\|^2_{L^2(\mathcal{S}, \ell^2)}\right)^{\frac{1}{3}}.
\end{equation*}
$\square$

\subsection{Discussion on Selection Consistency}

In the main text, we present the regret bound in order of $\mathcal{O}((\|r-r^*\|^2_{L^2(\mathcal{S}, \ell^2)})^{1/(3-2\alpha)})$.
It is often of interest to give the result of the selection consistency for a given reward estimator $r$.
\begin{lemma}
\label{consistency}
Given Assumption \ref{noisep} and the estimator from (\ref{eqn:MLE_sec3}), there exist an universal constant $c_6$ such that
\begin{equation*}
\mathbb{P}_{\mathcal{S}}\left(\pi_{r^*}(s)\neq\pi_{r}(s)\right) \leq c_6\left(\|r-r^*\|^2_{L^2(\mathcal{S}, \ell^2)}\right)^{\frac{\alpha}{3-2\alpha}},
\end{equation*}
 where $\pi_{r^*}(s)$ is an optimal policy that gives the optimal action that maximizes the reward. 
\end{lemma}
 This quantity is important in statistical machine learning literature \citep{Bartlett2006Surrogate,tsybakov2007fast}.
As $\alpha \rightarrow 1$, the selection consistency achieves the best rate.
When $\alpha \rightarrow 0$, following the Lemma \ref{noiser},  the reward of an alternative action is comparable with the optimal one; thus, the selection consistency result may fail.
Fundamentally, in reinforcement learning problems, we do not make predictions that are consistent with data labels. Still, it is a beneficial complement for us to understand the effects of the margin-type condition.

\subsection{Proof of Lemma \ref{consistency}}
By the definition of regret,
\begin{equation*}
\begin{aligned}
\mathcal{E}(r)=&\int_\mathcal{S}\left(r^*(s,\pi_{r^*}(s)) - r^*(s,\pi_{r}(s))\right)d \rho_s  \\ \geq& \int_\mathcal{S}\left(r^*(s,\pi_{r^*}(s)) - r^*(s,\pi_{r}(s))\right) \1\left\{r^*(s, \pi_{r^*}(s)) - \max_{a\in \mathcal{A}/\pi_{r^*}(s)}r^*_a(s)>t\right\} d \rho_s   \\
\geq& t\int_\mathcal{S} \1\left\{\pi_{r^*}(s)\neq\pi_{r}(s)\right\} \cdot \1\left\{r^*(s, \pi_{r^*}(s)) - \max_{a\in \mathcal{A}/\pi_{r^*}(s)}r^*_a(s)>t\right\}d \rho_s  \\
\geq & t\left(\mathbb{P}_{\mathcal{S}}\left(\pi_{r^*}(s)\neq\pi_{r}(s)\right)-c_g t^{\frac{\alpha}{1-\alpha}}\right).
\end{aligned}
\end{equation*}
The last line follows $\mathbb{P}(E_1 \cap E_2)\geq \mathbb{P}(E_1)-\mathbb{P}(E_2^c)$, for any two events $E_1$ and $E_2$.  Minimizing this term with respect to $t$, \textit{i.e.}, taking $t=c' \mathbb{P}_{\mathcal{S}}\left(\pi_{r^*}(s)\neq\pi_{r}(s)\right)^{(1-\alpha)/\alpha}$ results in

\begin{equation*}
    \mathbb{P}_{\mathcal{S}}\left(\pi_{r^*}(s)\neq\pi_{r}(s)\right) \leq \frac{1}{{c^{\prime}}^\alpha} \mathcal{E}(r)^\alpha
    \leq  c_6\left(\|r-r^*\|^2_{L^2(\mathcal{S}, \ell^2)}\right)^{\frac{\alpha}{3-2\alpha}},
\end{equation*}
where $c^\prime$ and $c_6$ are constants depending on $c_g$ and $\alpha$.
$\square$

\section{Proof of Error Bounds}
\subsection{Proof of Lemma \ref{est_errors}}
Denote $\tilde{r}$ as an estimator that maximizes the likelihood in the function class $\mathcal{F}_{\operatorname{DNN}}$ as
$$
\tilde{r}=\underset{r \in \mathcal{F}_{\operatorname{DNN}}}{\arg \max } \ l(r) .
$$
We expand the excess risk by adding and substituting the following terms:
$$
\begin{aligned}
l(r^*)-l\left(\hat{r}\right) = & \left[\hat{l}(\hat{r})-l(\hat{r})\right]+\left[\hat{l}\left(\tilde{r}\right)-\hat{l}(\hat{r})\right]+\left[l\left(\tilde{r}\right)-\hat{l}(\tilde{r})\right]+\left[l\left(r^*\right)-l\left(\tilde{r}\right)\right]  \\
\leq & {\left[l(\hat{r})-\hat{l}(\hat{r})\right]+\left[l\left(\tilde{r}\right)-\hat{l}\left(\tilde{r}\right)\right]+\left[l\left(r^*\right)-l\left(\tilde{r}\right)\right] } \\
\leq & 2 \sup _{r \in \mathcal{F}_{\operatorname{DNN}}}\left|l(r)-\hat{l}(r)\right|+l\left(r^*\right)-l\left(\tilde{r}\right) \\
= & 2 \sup _{r \in \mathcal{F}_{\operatorname{DNN}}}\left|l(r)-\hat{l}(r)\right|+\inf _{r \in \mathcal{F}_{\operatorname{DNN}}}\left[l(r^*)-l\left(r\right)\right],
\end{aligned}
$$
where the first inequality follows from the definition of $\hat{r}$ as the maximizer of $\hat{l}(r)$ in $\mathcal{F}_{\operatorname{DNN}}$, then $\hat{l}\left(\tilde{r}\right)-\hat{l}(\hat{r}) \leq 0$. The second inequality holds due to the fact that both $\hat{r}$ and $\tilde{r}$ belong to the function class $\mathcal{F}_{\operatorname{DNN}}$, and the last equality is valid by the definition of $\tilde{r}$.

\subsection{Proof of Proposition \ref{stochastic_error}}

Let $Z_{i}(r):=\log g\left(y^i ; r(s^i,a_1^i)-r(s^i,a_0^i)\right)$ for $ 1\leq i \leq N.$ Then, given Assumption \ref{dynamic_range}
and Hoeffding’s inequality, with probability at least $1-\delta$, we have
$$\left|\frac{1}{N}\sum_{i=1}^N\left(Z_{i}(r)-\mathbb{E}\left[Z_i(r)\right]\right)\right|\leq \kappa_0\sqrt{\frac{\log\left(\frac{2}{\delta}\right)}{2N}},$$
where the expectation $\mathbb{E}$ is taken over the data distribution. Now considering an estimator $r$ that is within the deep neural network function class, We can further obtain that for any given $\tau>0$, let $r_1, r_2, \ldots, r_{\mathcal{N}}$ be the anchor points of an $\tau$-covering for the function class $\mathcal{F}_{\operatorname{DNN}}$, where we denote 
$\mathcal{N}:=
\mathcal{N}\left(\mathcal{F}_{\operatorname{DNN}},\tau,\|\cdot\|_{L^\infty\left(\mathcal{S}, \ell^\infty\right)}\right)$
as the covering number 
 of $\mathcal{F}_{\operatorname{DNN}}$ with radius $\tau$ under the norm $\|\cdot\|_{L^\infty\left(\mathcal{S}, \ell^\infty\right)}$. By definition, for any $r \in \mathcal{F}_{\operatorname{DNN}}$, there exists an anchor $r_h$ for $h \in\{1, \ldots, \mathcal{N}\}$ such that $\left\|r_h-r\right\|_{L^\infty\left(\mathcal{S}, \ell^\infty\right)} \leq \tau$. We further decompose the stochastic error as follows
$$
\begin{aligned}
l(r)-\hat{l}(r) \leq & l\left(r_h\right)-\hat{l}\left(r_h\right)+l(r)-l\left(r_h\right)+\hat{l}\left(r_h\right)-\hat{l}(r) \\
= & l\left(r_h\right)-\hat{l}\left(r_h\right) +\frac{1}{N}\left|\sum_{i=1}^N\mathbb{E}\left[Z_{i}(r)-Z_{i}(r_h)\right]\right|+\frac{1}{N}\left|\sum_{i=1}^N\left[Z_{i}(r)-Z_{i}(r_h)\right]\right| \\
\leq & l(r_h)-\hat{l}(r_h)+2\kappa_1\tau,
\end{aligned}
$$
\noindent where the last step is due to the Lipschitz property of $\log g(y,u)$. Therefore, with a fixed $\epsilon>0$,
\begin{equation}
\label{covering}
\begin{aligned}
& \mathbb{P}\left(\sup _{r \in \mathcal{F}_{\operatorname{DNN}}}\left|l(r)-\hat{l}(r)\right| \geq (\epsilon+\kappa_1\tau)\right) \leq \mathbb{P}\left(\exists \ h \in\{1, \ldots, \mathcal{N}\}:\left|l\left(r_h\right)-\hat{l}\left(r_h\right)\right| \geq \epsilon\right) \\
&\leq \mathcal{N}_n\left(\mathcal{F}_{\operatorname{DNN}},\tau,\|\cdot\|_{L^\infty\left(\mathcal{S}, \ell^\infty\right)}\right) \max _{h \in\{1, \ldots, \mathcal{N}\}} \mathbb{P}\left(\left|l \left(r_h\right)-\hat{l} \left(r_h\right)\right| \geq \epsilon\right) \\
&\leq 2 
\mathcal{N}_n
\left(\mathcal{F}_{\operatorname{DNN}},\tau, \|\cdot\|_{L^\infty
\left(\mathcal{S}, \ell^\infty\right)}
\right) \exp \left(-\frac{2N\epsilon^2}{\kappa_0^2}  \right),
\end{aligned}
\end{equation}
where the last line comes from Hoeffding’s inequality. Then, for any $\delta>0$, let $\tau=1 /N$ and $\epsilon=\kappa_0 \sqrt{2\log \left(2 \mathcal{N}\left( \mathcal{F}_{\operatorname{DNN}},\tau,\|\cdot\|_{L^\infty\left(\mathcal{S}, \ell^\infty\right)}\right) / \delta\right) / N}$ so that the right-hand side of (\ref{covering}) equals to $\delta$, we have
\begin{equation*}
    \mathbb{P}\left(\sup_{r \in \mathcal{F}_{\operatorname{DNN}}}\left|l(r)-\hat{l}(r)\right| \geq \epsilon+2\kappa_1 \tau\right) \leq \delta.
\end{equation*}
In other words, with probability at least $1-\delta$,
\begin{equation}
\begin{aligned}
& \sup _{r \in \mathcal{F}_{\operatorname{DNN}}}\left|l(r)-\hat{l}(r)\right| \leq \epsilon+2\kappa_1 \tau \\
& \leq \sqrt{2}\kappa_0  \left(\sqrt{\frac{\log 2\sqrt{2} \mathcal{N}\left(\tau, \mathcal{F}_{\operatorname{DNN}},\|\cdot\|_{L^\infty
\left(\mathcal{S}, \ell^\infty\right)}\right)}{N}}   +   \sqrt{\frac{\log \left(\frac{1}{\delta}\right)}{N}}  \right)+\frac{2\kappa_1}{N},
\end{aligned}
\label{boundwithcovering}
\end{equation}
where in the second step we use the inequality $\sqrt{C_1+C_2} \leq \sqrt{C_1}+\sqrt{C_2}$, for any $C_1, C_2 \geq 0$.

In the rest of the proof, we bound the covering number. 
Without loss of generality, we also define classes of sub-networks of $\mathcal{F}_{\operatorname{DNN}}$, that is, $\{\mathcal{F}_{1},\mathcal{F}_{2},\cdots,\mathcal{F}_{|\mathcal{A}|}\}$, with non-sharing hidden layers. By doing so, the function $r_a \in \mathcal{F}_{a}$ in each reduced function class takes states as input and returns $r(s, a)$ given an action $a\in \mathcal{A}$. For convenience, we assume all the sub-networks have the same width in each layer, and all model parameters are bounded within $[-1,1]$, that is,
\begin{align*}
    \mathcal{F}_a\left(W,D\right)=\left\{r_a(\cdot ; \theta): \mathbb{R}^{d} \rightarrow \mathbb{R} \text { defined in (\ref{neural}) }: \theta \in[-1, 1]^p \right\},
\end{align*}
such that the function class of our interest is covered by the product space of sub-network classes, that is, $\mathcal{F}_{\operatorname{DNN}} \subset \mathcal{F}_{1}\otimes\mathcal{F}_{2}\otimes\cdots\otimes\mathcal{F}_{|\mathcal{A}|}$. 
Now, we can express the covering number in terms of the product of complexities of sub-networks by \begin{equation}
\label{relation_covering_number}
    \log \mathcal{N}\left(\mathcal{F}_{\operatorname{DNN}}, \tau,\|\cdot\|_{L^\infty\left(\mathcal{S}, \ell^\infty\right)}\right) \leq |\mathcal{A}| \log \mathcal{N}(\mathcal{F}_a, \tau/|\mathcal{A}|,\|\cdot\|_{L^{\infty}}).
\end{equation} For the deep ReLU neural network in our setting, \citet[Theorem 2]{shen2024complexity} shows that for any $\tau>0$,
$$
\mathcal{N}\left(\mathcal{F}, \tau,\|\cdot\|_{\mathcal{L}_\infty}\right) \leq \frac{\left(2^{D+5}(D+1)W^D \cdot \tau^{-1}\right)^p}{(W!)^D}.
$$ 
Then, apply the above inequality to  (\ref{relation_covering_number}),
\begin{equation}
    \log\mathcal{N}\left(\mathcal{F}_{\operatorname{DNN}}, \tau,\|\cdot\|_{L^\infty\left(\mathcal{S}, \ell^\infty\right)}\right) \leq 
    |\mathcal{A}| Dp \log \left(\frac{2 W (32(D+1)|\mathcal{A}|/\tau)^{1/D}}{(W!)^{1/p}}\right).
\label{coveringnumber}
\end{equation}
Plug (\ref{coveringnumber}) in (\ref{boundwithcovering}), and take $\tau=1 / N$, with probability at least $1-\delta$,
\begin{equation*}
\begin{aligned}
& \sup _{r \in \mathcal{F}_{\operatorname{DNN}}}\left|l(r)-\hat{l}(r)\right|\\
& \leq \kappa_0\sqrt{\frac{2}{N}} \left(\sqrt{\log 2\sqrt{2} |\mathcal{A}| Dp \log \left(\frac{2 W (32(D+1)|\mathcal{A}|N)^{1/D}}{(W!)^{1/p}}\right)}   +   \sqrt{\log (1/\delta)}  \right)+\frac{2\kappa_1}{N}\\
& \leq \kappa_0\sqrt{\frac{2}{N}} \left(c_3\sqrt{|\mathcal{A}| Dp \log \left(\frac{ W ((D+1)|\mathcal{A}|N)^{1/D}}{(W!)^{1/p}}\right)}   +   \sqrt{\log (1/\delta)}  \right).
\end{aligned}
\end{equation*}
In the second step, since $1/N$ is decaying faster than $\sqrt{\log (N)/N}$, we can simplify the expression by making $c_3$ a universal constant.
$\square$

\subsection{Proof of Proposition \ref{approx_error}}
We adopt the ReLU network approximation result for H\"older smooth functions in $\mathcal{H}^\beta([0,1]^d, c_{\mathcal{H}})$, proposed in \citet[Theorem 3.3]{jiao2023deep}. With Assumption \ref{Holder_assumption}, for any $M_1, M_2 \in \mathbb{N}^{+}$, and for each $a \in \mathcal{A}$, there exists a function $\tilde{r}$ implemented by a ReLU network with width $W=38(\lfloor\beta\rfloor+1)^2 d^{\lfloor\beta\rfloor+1} M_1\lceil\log _2(8 M_1)\rceil$ and depth $D=21(\lfloor\beta\rfloor+1)^2 M_2\lceil\log _2(8 M_2)\rceil$ such that
$$
|\tilde{r}(s,a)-r^*(s,a)| \leq 18 c_{\mathcal{H}}(\lfloor\beta\rfloor+1)^2 d^{\lfloor\beta\rfloor+(\beta \vee 1) / 2}(M_1 M_2)^{-2 \beta / d},
$$
for all $s \in[0,1]^d$ except a small set $\Omega \in \mathcal{S}$ with Lebesgue measure $\delta K p$ for any $\delta >0$. 
Since we have the same setting for all sub-networks, we have the same bound for $\max_{a\in\mathcal{A}}|\tilde{r}(s,a)-r^*(s,a)|$. Therefore, integrating both sides with respect to the state space distribution, we have
{\small
\begin{equation}
\label{except_delta_measure}
\begin{aligned}
    \|\tilde{r}-r^*\|_{L^1(\mathcal{S}, \ell^\infty)} &\leq 18 c_{\mathcal{H}}(\lfloor\beta\rfloor+1)^2 p^{\lfloor\beta\rfloor+(\beta \vee 1) / 2}(M_1 M_2)^{-2 \beta / d}+\mathbb{P}(\Omega) \cdot \sup _{s \in \Omega}\{\max_{a\in\mathcal{A}}|\tilde{r}(s,a)-r^*(s,a)|\}. 
\end{aligned}
\end{equation}}
By Assumption \ref{Holder_assumption}, the marginal distribution of $\mathcal{S}$ is absolutely continuous with respect to the Lebesgue measure, which means that $\liminf _{\delta \rightarrow 0} \mathbb{P}(\Omega)=0$. Meanwhile, we know from the definition of $\mathcal{F}_{\operatorname{DNN}}$ and Assumption \ref{dynamic_range} that both $\sup_{s\in\mathcal{S}}|\tilde{r}(s,a)|$ and $\sup_{s\in\mathcal{S}}|r^*(s,a)|$ are bounded for all $a \in \mathcal{A}$. Therefore, by taking the limit infimum with respect to $\delta$ on both sides of (\ref{except_delta_measure}), we have
\begin{equation*}\|\tilde{r}-r^*\|_{L^1(\mathcal{S}, \ell^\infty)} \leq 18 c_{\mathcal{H}}(\lfloor\beta\rfloor+1)^2 d^{\lfloor\beta\rfloor+(\beta \vee 1) / 2}(M_1 M_2)^{-2 \beta / d} .
\end{equation*}
Therefore, by the Lipschitz property of the likelihood function,
\begin{equation*}l(r^*)-l(\tilde{r})\leq 2\kappa_1\left\|\tilde{r}-r^*\right\|_{L^1(\mathcal{S}, \ell^\infty)}\leq36\kappa_1 c_{\mathcal{H}}(\lfloor\beta\rfloor+1)^2 d^{\lfloor\beta\rfloor+(\beta \vee 1) / 2}(M_1 M_2)^{-2 \beta / d}.
\end{equation*}
$\square$

\subsection{Proof of Theorem \ref{non_asymp_error}}
Before dealing with the error terms, we first control the logarithmic term involved in the covering number. Let $c$ be a positive universal constant, then
\begin{equation*}
\begin{aligned}
\frac{W((D+1)|\mathcal{A}| N)^{1 / D}}{(W!)^{1 / p}} &\leq 
\frac{W((D+1)|\mathcal{A}| N)^{1 / D}}{\left(e\left(W/e\right)^W\right)^{1 / p}} \\
&= c\left(\frac{W((D+1)|\mathcal{A}| N)^{1 / D}}{W^{1 / \mathcal{W D}}}\right) \\
&= c\left(W(|\mathcal{A}|N)^{1 / D}\right).
\end{aligned}
\end{equation*}
The first step is from the inequality that $n!\geq e(n/e)^n$, for any $n \in \mathbb{N^+}$. The second step holds for rectangular neural networks, where the size $ p = \mathcal{O}(W^2 D)$. The last step follows Proposition \ref{approx_error}, since we require the network has width $W>114$ and depth $D>63$, leading that $W^{1/W} \asymp 1$ and $(D+1)^{1/D} \asymp 1$. Now, we combine both stochastic error and approximation error.
\begin{equation*}
\begin{aligned}
    l(\hat{r})-l(r^*) &\leq 2\kappa_0\sqrt{\frac{2}{N}}  \left(c_3\sqrt{ |\mathcal{A}| Dp \log \left(c\left(W (|\mathcal{A}|N)^{1/D}\right)\right)}   +   \sqrt{\log (1/\delta)}  \right)\\
    &\quad\quad\quad\qquad\qquad\qquad\qquad\qquad+36\kappa_1 c_{\mathcal{H}}(\lfloor\beta\rfloor+1)^2 d^{\lfloor\beta\rfloor+(\beta \vee 1) / 2}(M_1 M_2)^{-2 \beta / d}.
\end{aligned}
\end{equation*}
To balance these two error terms, proper tuning parameters $M_1$ and $M_2$ are selected to optimize the convergence rate with the most efficient network design. 
Since the network size grows linearly in depth $D$ but quadratically in width $W$. To reach the optimal rate while, at the same time, saving network size, we decide to fix $M_1=1$ so that the width $W$ is growing with a polynomial of the input dimension $d$ while independent of sample size $N$. Meanwhile, we take $M_2=\left\lfloor N^{d/(2d+4\beta)}\right\rfloor$. Then with $d\gg \beta$, we have
\begin{equation}
\begin{gathered}
\label{weight_and_depth}
W=114(\lfloor\beta\rfloor+1)^2 d^{\lfloor\beta\rfloor+1};
D=21(\lfloor\beta\rfloor+1)^2\left\lceil N^{\frac{d}{2d+4\beta}} \log _2\left(8 N^{\frac{d}{2d+4\beta}}\right)\right\rceil =\mathcal{O}\left(\sqrt{N}\right),
\end{gathered}
\end{equation}
and
$$p =\mathcal{O}\left(W^2 D\right)=\mathcal{O}\left((\lfloor\beta\rfloor+1)^6 d^{2\lfloor\beta\rfloor+2}\left\lceil N^{\frac{d}{ 2(d+2 \beta)}}\left(\log _2 N\right)\right\rceil\right)=\mathcal{O}\left(\sqrt{N}\right),$$
where the $\log(N)$ factors are omitted for simplicity.

Therefore, combining (\ref{weight_and_depth}) with the error bound, and letting $c^{\prime}_3$ an another universal constant, with probability at least $1-\delta$,
\begin{equation*}
\begin{aligned}
    l(\hat{r})-l(r^*) \leq & 2\kappa_0\sqrt{\frac{2}{N}}  \left(c^{\prime}_3\sqrt{ |\mathcal{A}|}(\lfloor\beta\rfloor+1)^4 d^{\lfloor\beta\rfloor+1}(\log (N))^2 N^{\frac{d}{2d+4\beta}}  +   \sqrt{\log (1/\delta)}  \right)\\
    &\quad\quad\quad\quad\quad\quad\quad\quad\quad\quad\quad\quad\quad\quad\quad\quad\quad+36\kappa_1  c_{\mathcal{H}}(\lfloor\beta\rfloor+1)^2 d^{\lfloor\beta\rfloor+(\beta \vee 1) / 2} N^{\frac{-\beta}{d+2\beta}}\\
    \leq & 2\sqrt{2}\kappa_0  \left(c_4\sqrt{ |\mathcal{A}|}(\lfloor\beta\rfloor+1)^4 d^{\lfloor\beta\rfloor+1}(\log (N))^2 N^{\frac{-\beta}{d+2\beta}}  +   \sqrt{\frac{\log (1/\delta)}{N}}  \right).
\end{aligned}
\end{equation*}
Since the stochastic error and approximation error are of the same order now, in the last step, we combine them with $c_4$, another universal constant. In this part, we find the following connection between the excess risk $l(\hat{r})-l(r^*)$ and the estimation error $\|\hat{r}-r^*\|^2_{L^2(\mathcal{S}, \ell^2)}$.
The first-order optimality condition of $l(\cdot)$ implies $\nabla l(r^*)=0$, then
\begin{equation*}
    \begin{aligned}
        l(r^*)-l(\hat{r}) &= \nabla l(r^*) (r^*-\hat{r})+\frac{1}{2}(r^*-\hat{r})^\top \nabla^2 l(\zeta)(r^*-\hat{r}) \\
        &= 0 + \int_{\mathcal{S}}\frac{1}{N}\sum_{i<j} n_{ij}\frac{\partial^2}{\partial u^2} \log g(y ; \zeta)(\hat{r}(s,a_i)-\hat{r}(s,a_j)- (r^*(s,a_i)-r^*(s,a_j)))^2 d\rho_s\\
        & \geq  \int_{\mathcal{S}}\sum_{i<j} \frac{n_{ij}}{N} \kappa_2 (\hat{r}(s,a_i)-\hat{r}(s,a_j)- (r^*(s,a_i)-r^*(s,a_j)))^2 d\rho_s\\
        &\geq \kappa_2\int_{\mathcal{S}}\kappa_{\Lambda}\sum_{a\in \mathcal{A}}\left(\hat{r}(s,a)-r^*(s,a)\right)^2 d\rho_s\\
        & = \kappa_2 \kappa_{\Lambda}\|\hat{r}-r^*\|^2_{L^2(\mathcal{S}, \ell^2)},
    \end{aligned}
\end{equation*}
where there exists $y \in \Omega$ and $ \zeta \in [\hat{r}(s,a_i)-\hat{r}(s,a_j), r^*(s,a_i)-r^*(s,a_j)]$. 

The second last step holds by the identifiability constraint for both the true and estimated reward function, i.e.,  $\sum_{a\in\mathcal{A}}\hat{r}(s,a) = 0$ and $\sum_{a\in\mathcal{A}}r^*(s,a) = 0$, which leads to  $\sum_{a\in \mathcal{A}} (\hat{r}(s,a) - r^*(s,a))=0$. This identifiability constraint is also posited in \cite{shah2016estimation}. 
Therefore, we conclude that the final non-asymptotic error bound: with probability at least $1-\delta$,
\begin{equation*}
    \|\hat{r}-r^*\|^2_{L^2(\mathcal{S}, \ell^2)} \leq 2\sqrt{2}\frac{\kappa_0}{\kappa_2\kappa_{\Lambda}} \left(c_4\sqrt{ |\mathcal{A}|}(\lfloor\beta\rfloor+1)^4 d^{\lfloor\beta\rfloor+1}(\log (N))^2 N^{-\frac{\beta}{d+2\beta}}  +   \sqrt{\frac{\log (1/\delta)}{N}}  \right).
\end{equation*}
$\square$

Different from the deep regression and classification problem, the convergence of the excess risk does not directly ensure the functional convergence of the estimated reward.
It is necessary to consider additional constraints on the dataset structure (see Assumption \ref{assump_datacover}).
Assumption \ref{assump_datacover} ensures the  Laplacian spectrum of the comparison graph to be bounded away from 0 and ensures the convergence of the estimator $\hat r$.
The condition is also considered by \citet{zhu2023principled}. 
We note that to the best of our knowledge, we are the first to study its role in the non-parametric analysis for deep reward modeling.

We would like to comment that our theoretical results align with the scaling law, which suggests how to decide network size as the sample size increases \citep{havrilla2024understanding}. This result arises from analyzing the approximation capacity and complexity (training efficiency) of networks. As the complexity of the network increases, it gains greater approximation power leading to a smaller approximation error for representing the target function, whereas the stochastic error increases as more parameters need to be trained or estimated. Our theory is designed to balance this trade-off. Specifically, when the sample size is sufficiently large, stochastic error becomes negligible, allowing the approximation error to dominate. Under such cases, large-scale complex network structures are advantageous, and performance can be enhanced with increased model scale \citep{jiao2023deep,havrilla2024understanding}.

\section{Additional Experimental Results}
\label{appendix_experiment}
All experiments involved in this work are conducted on a Linux server equipped with AMD EPYC 7763 64-Core Processor with 1.12 TB RAM.

\begin{figure}[!ht]
    \centering
    \includegraphics[width=0.9\linewidth]{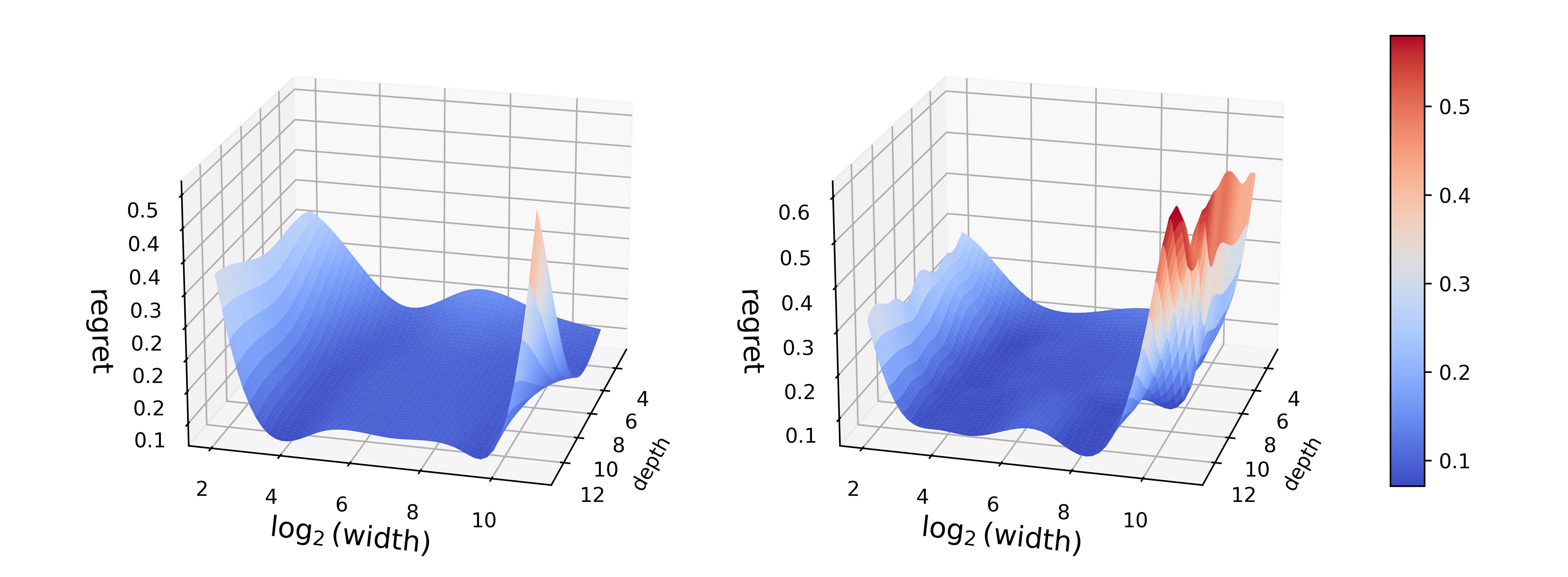}
    \caption{Surface plot of regret for both BT (left panel) and Thurstonian (right panel) models for Hermite-Gaussian reward functions across different neural network configurations.}
    \label{fig:HG_two}
\end{figure}

\begin{figure}[!ht]
    \centering
    \includegraphics[width=0.9\linewidth]{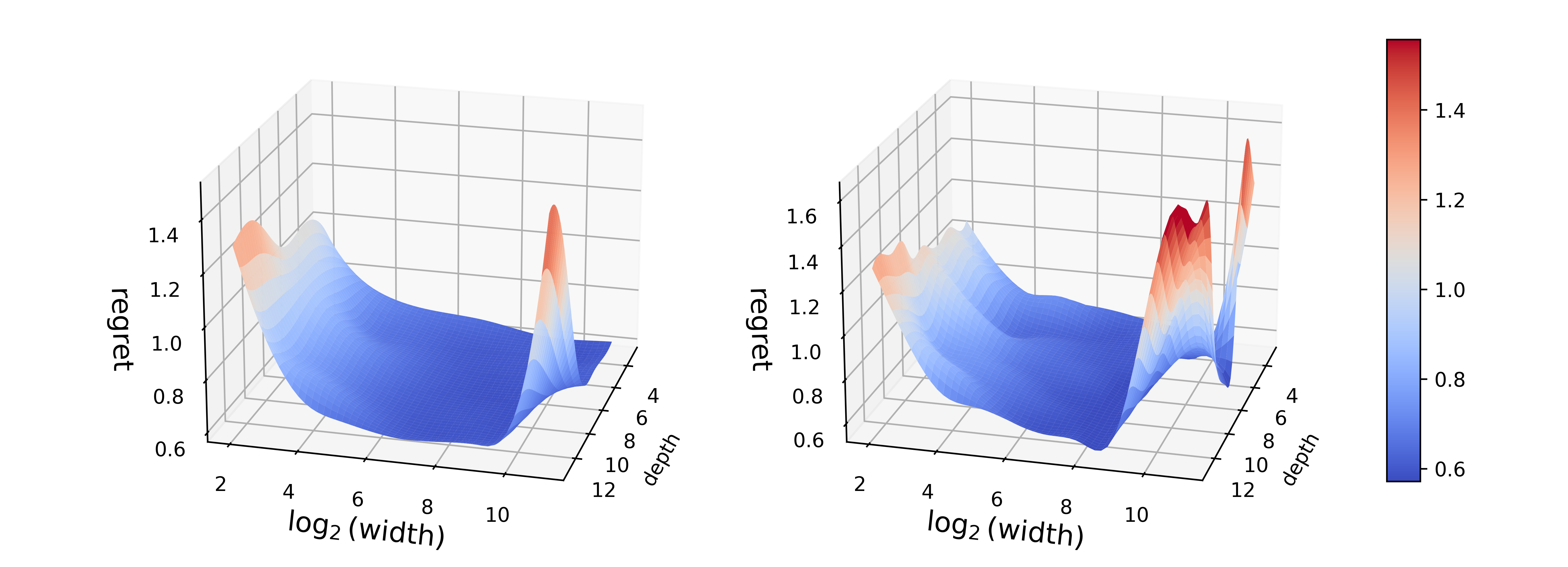}
    \caption{Surface plot of regret for both BT (left panel) and Thurstonian (right panel) models for nonlinear composite sinusoid reward functions across different neural network configurations.}
    \label{fig:Ss_two}
\end{figure}

\begin{figure}[!ht]
    \centering
    \begin{subfigure}{}
        \centering
        \includegraphics[width=0.4\textwidth]{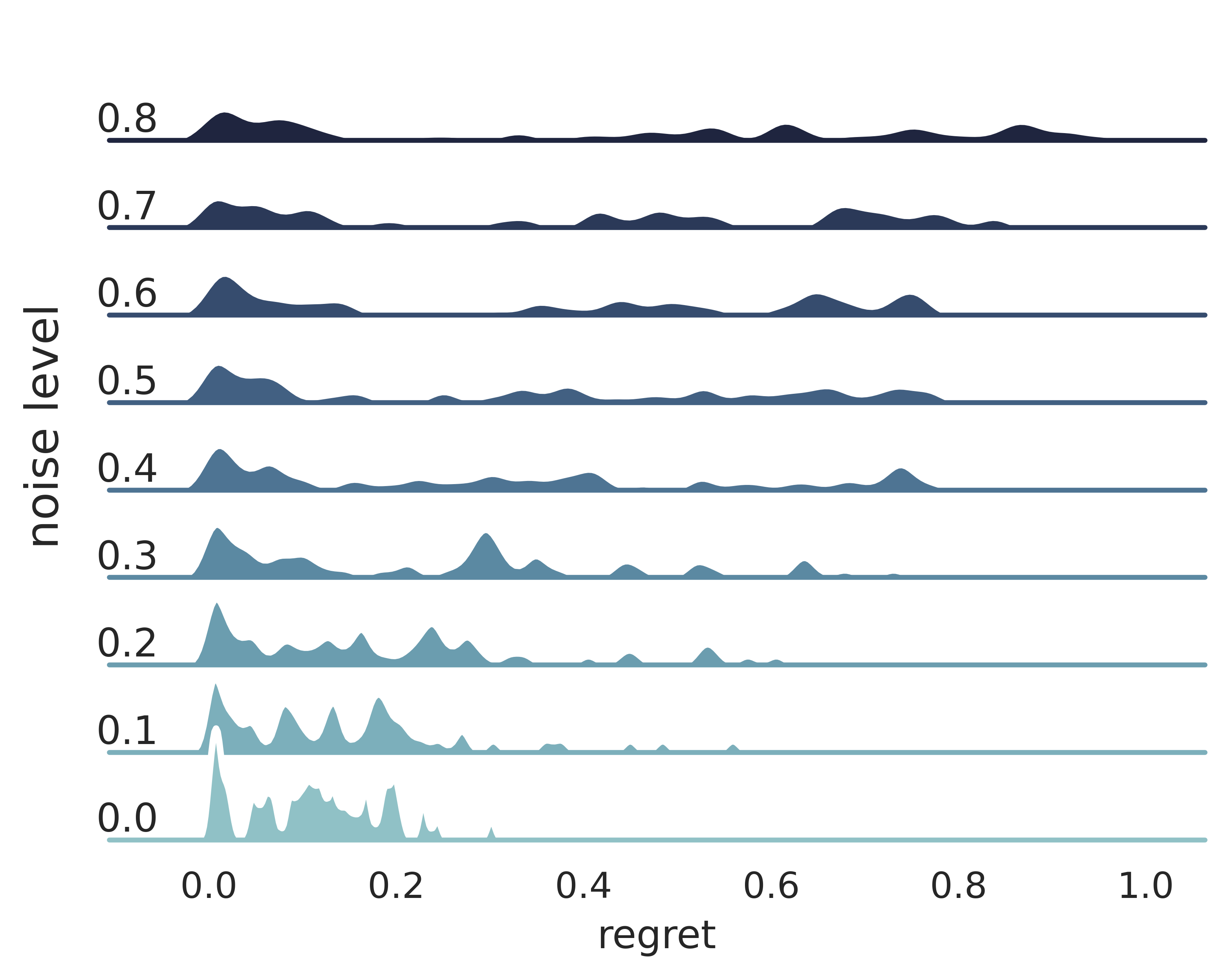}  
    \end{subfigure}
    \begin{subfigure}{}
        \centering
        \includegraphics[width=0.4\textwidth]{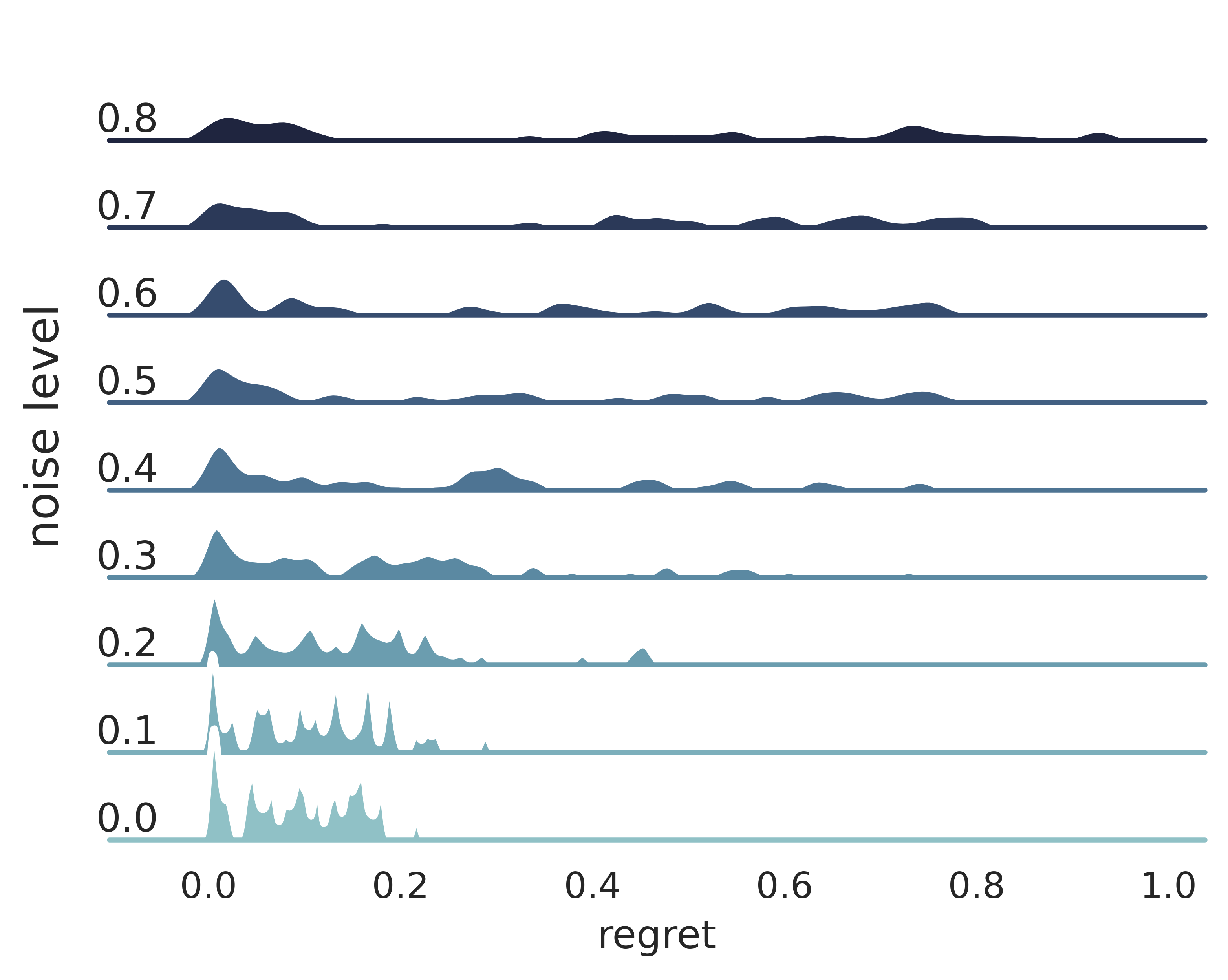}  
    \end{subfigure}
    \caption{Empirical distribution of regret for both BT (left panel) and Thurstonian (right panel) models for Hermite-Gaussian reward functions under different noise levels. }
    \label{fig:Hg_margin}
\end{figure}

\begin{figure}[!ht]
    \centering
    \begin{subfigure}{}
        \centering
        \includegraphics[width=0.4\textwidth]{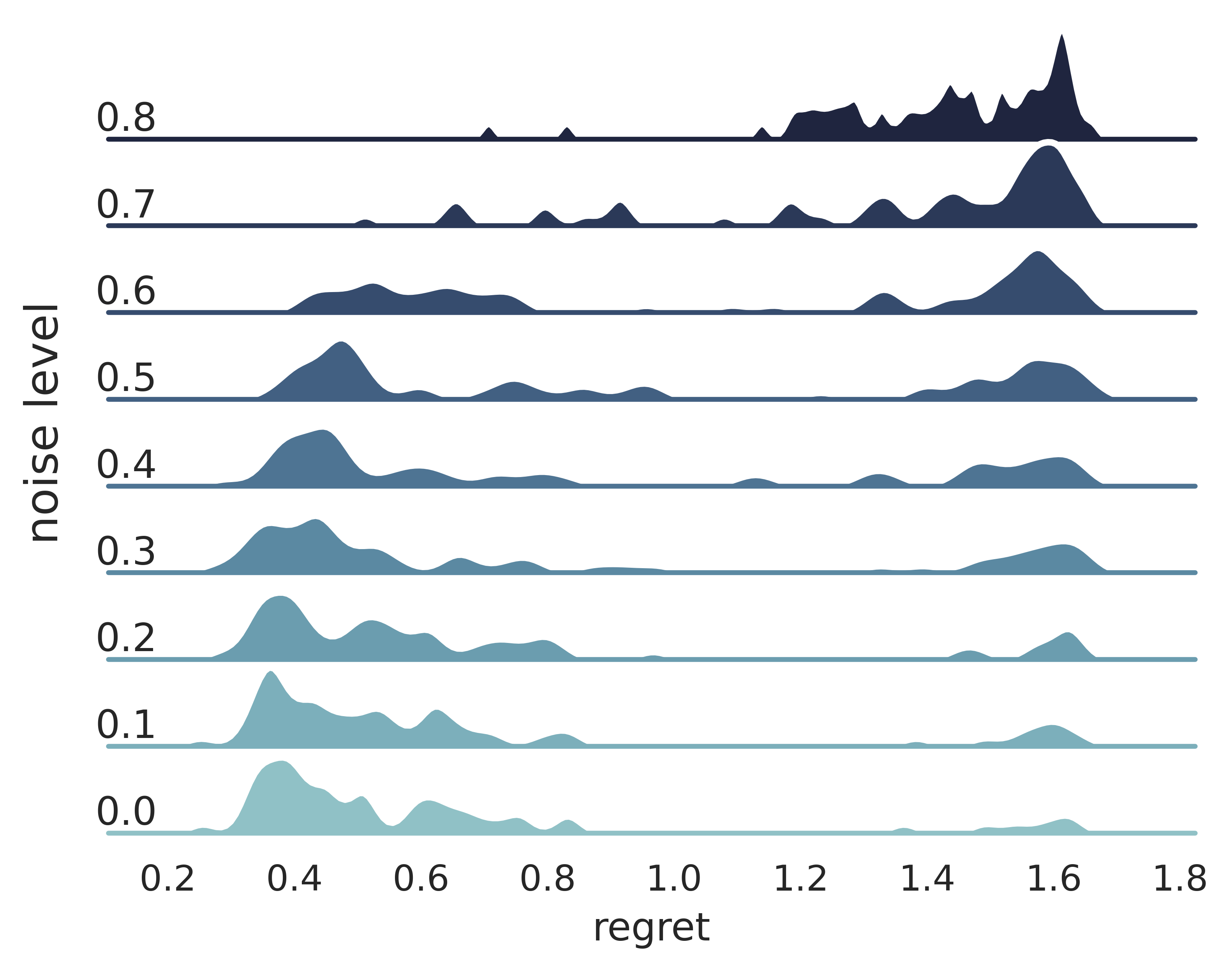}  
    \end{subfigure}
    \begin{subfigure}{}
        \centering
        \includegraphics[width=0.4\textwidth]{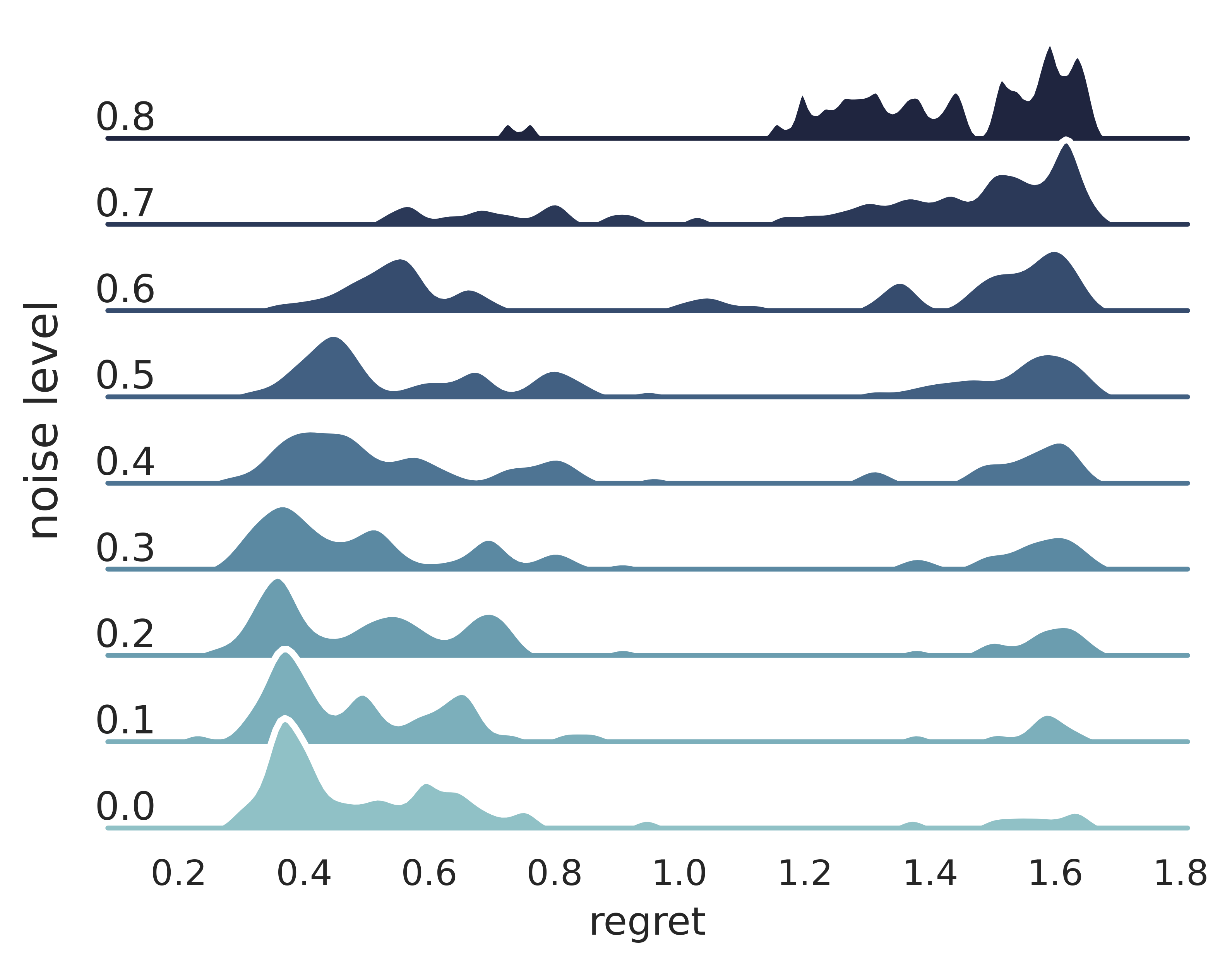}  
    \end{subfigure}
    \caption{Empirical distribution of regret for both BT (left panel) and Thurstonian (right panel) models for nonlinear composite sinusoid reward functions under different noise levels.}
    \label{fig:Ss_margin}
\end{figure}

To further validate the robustness of our theoretical findings, we extend the experiments in Section \ref{sec:exp} by evaluating two additional types of reward functions under the same protocol. These functions introduce more complex nonlinearities, testing the generalizability of our margin-based regret bounds. 
We retain the core setup from Section \ref{sec:exp} but 
the reward functions are thus defined as $r^*(s,a_1) = 2\psi(4\phi(s)^\top w^*)$ and $r^*(s,a_0) = -r^*(s,a_1)$, with $\phi(s) = (\sin(s_1), \ldots, \sin(s_d))^\top$ and $w^*$ is sampled from a standard normal distribution. We consider two types of the function $\psi(x)$ as follows,
\begin{itemize}
    \item Hermite-Gaussian function: 
   \begin{equation*}  
   \psi(x) = \left( \sqrt{15} \pi^{\frac{1}{4}}\right)^{-1}\left(4 x^5 - 20 x^3 + 15 x\right) e^{-\frac{1}{2} x^2}.
   \end{equation*}  
   \item Nonlinear composite sinusoidal function:  
   \begin{equation*}  
   \psi(x) = \sin(x) + \sin(x^2).
   \end{equation*}  
\end{itemize}
The results are consistent with our theoretical predictions and the experiment results presented in the main text, where the larger networks initially reduce regret but eventually degrade performance due to inflated stochastic error (Figures \ref{fig:HG_two} and \ref{fig:Ss_two}). A higher noise level implies a lower quality of the comparison dataset, resulting in higher regret and worse performance. (Figures \ref{fig:Hg_margin} and \ref{fig:Ss_margin}). These results solidify our claims that the margin condition and architecture-dependent bounds generalize to a broad reward function class. 

\FloatBarrier

\subsection{Additional Figures}
\label{appendix: additional}
\begin{figure}[!ht]
    \centering
    \begin{subfigure}{}
        \centering
        \includegraphics[width=0.48\textwidth]{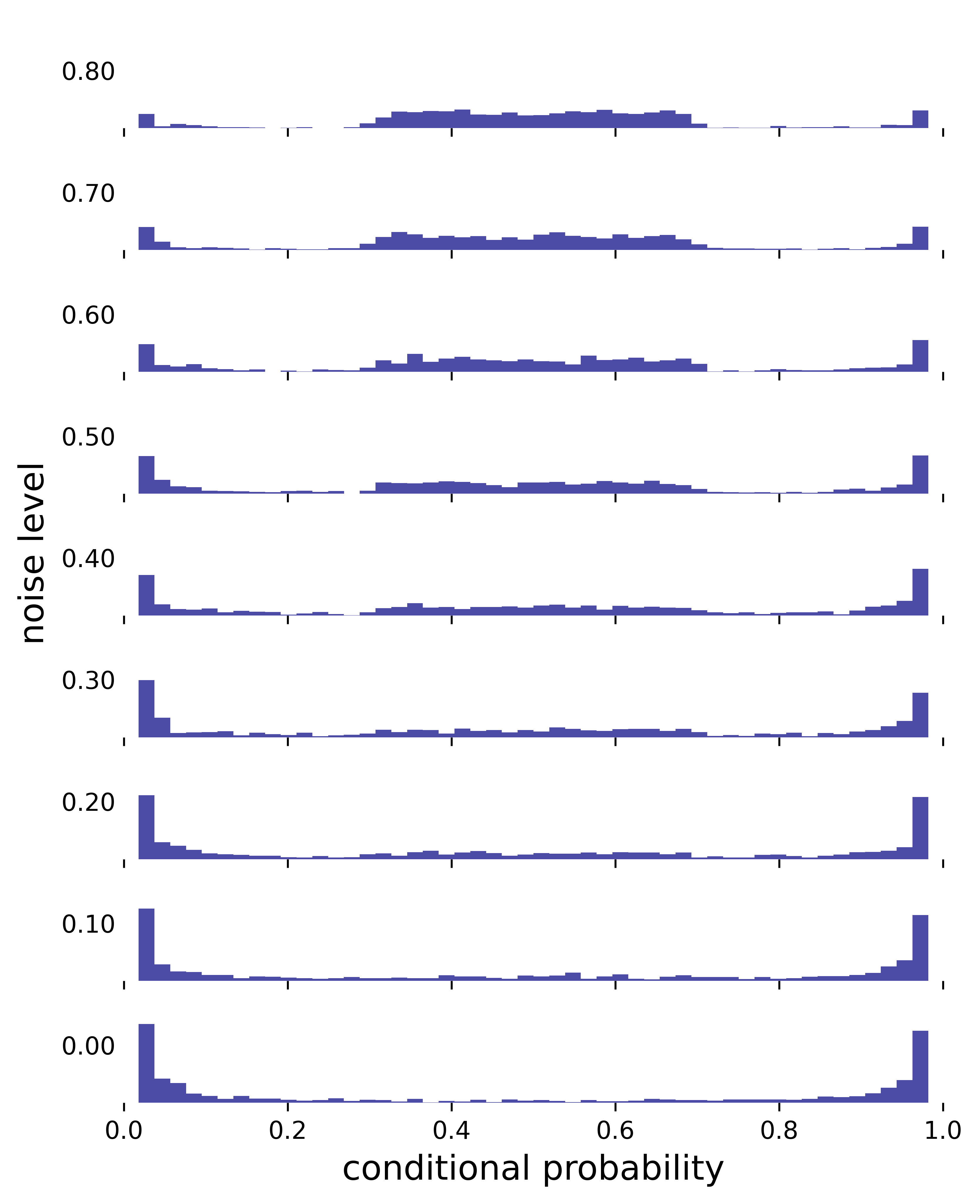}  
    \end{subfigure}
    \begin{subfigure}{}
        \centering
        \includegraphics[width=0.48\textwidth]{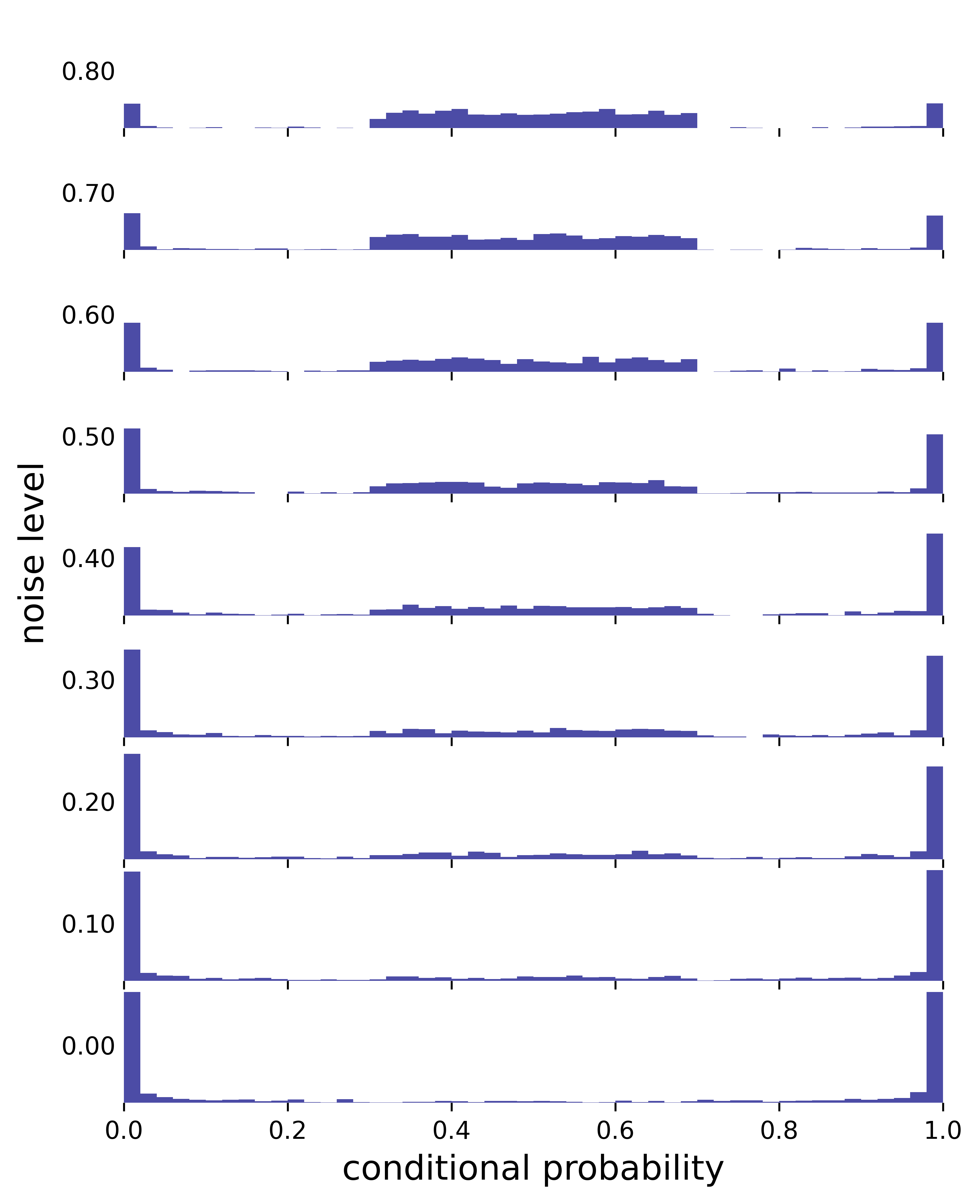}  
    \end{subfigure}
    \caption{
    Histogram of the calibrated condition probabilities $\mathbb{P}(y>0 \mid s,a_1,a_0)$  under different noise levels for both BT (left) and Thurstonian (right) models for sinusoid reward functions. As the noise level $m$ increases, the conditional probabilities exhibit denser dispersion around 0.5, indicating degradation in the comparison dataset quality and leading to increased regret. On the contrary, a small $m$ implies a higher quality of the comparison dataset with clearer human beliefs.}
    \label{fig:conditioal_probabilities}
\end{figure}  

\end{document}